\documentclass[runningheads]{llncs}

\usepackage{eccv}

\usepackage{mathtools}
\usepackage{caption} 
\usepackage{xspace}
\usepackage{multirow} 
\newcommand{\ourmethod}{\textsc{DriveVA}\xspace}
\usepackage{pifont}
\DeclareCaptionFont{mytiny}{\fontsize{6.5pt}{7.5pt}\selectfont}
\usepackage{xcolor}
\usepackage{pifont}
\usepackage{wrapfig}
\definecolor{gtfuture}{RGB}{44,160,44}
\definecolor{predfuture}{RGB}{214,39,40}
\definecolor{dpvogat}{RGB}{37,99,235}
\definecolor{dpvopred}{RGB}{245,158,11}

\newcommand{\cmark}{\textcolor{green!70!black}{\ding{51}}} 
\newcommand{\xmark}{\textcolor{red!70!black}{\ding{55}}}   

\usepackage{eccvabbrv}
\usepackage{enumitem}

\usepackage{graphicx}
\usepackage{booktabs}
\newcommand{\citep}[1]{\cite{#1}}
\newcommand{\citet}[1]{\cite{#1}}
\usepackage{float}

\usepackage[table]{xcolor}

\usepackage{amssymb}
\usepackage[accsupp]{axessibility}  

\usepackage[breaklinks,colorlinks,citecolor=eccvblue]{hyperref}

\usepackage{orcidlink}

\begin{document}

\title{\ourmethod: Video Action Models \\ are Zero-Shot Drivers}
\titlerunning{DriveVA: Video Action Models are Zero-Shot Drivers}

\makeatletter
\renewcommand{\@fnsymbol}[1]{\ensuremath{%
  \ifcase#1\or
    \ddagger\or
    \dagger\or
    \mathsection\or
    \mathparagraph\or
    \|\or
    **\or
    \dagger\dagger\or
    \ddagger\ddagger
  \else
    \@ctrerr
  \fi}}
\makeatother

\author{
Mengmeng Liu\inst{1} \and
Diankun Zhang\inst{2} \and
Jiuming Liu\inst{3} \and
Jianfeng Cui\inst{2} \and \\
Hongwei Xie\inst{2} \and
Guang Chen\inst{2} \and
Hangjun Ye\inst{2} \and \\
Michael Ying Yang\inst{4}\orcidlink{0000-0002-0649-9987} \and 
Francesco Nex\inst{1} \and
Hao Cheng\inst{1}\orcidlink{0000-0002-3254-4796}\thanks{Corresponding author.}
}

\authorrunning{M. Liu et al.}

\institute{
\makebox[\textwidth][c]{\footnotesize
\textsuperscript{1}University of Twente, The Netherlands\hspace{0.4em}
\textsuperscript{2}Xiaomi EV, China}\
\makebox[\textwidth][c]{\footnotesize
\textsuperscript{3}University of Cambridge, United Kingdom\hspace{0.4em}
\textsuperscript{4}University of Bath, United Kingdom}
}
\maketitle

\begin{figure}[t]
    \centering
    \includegraphics[width=1.0\linewidth]{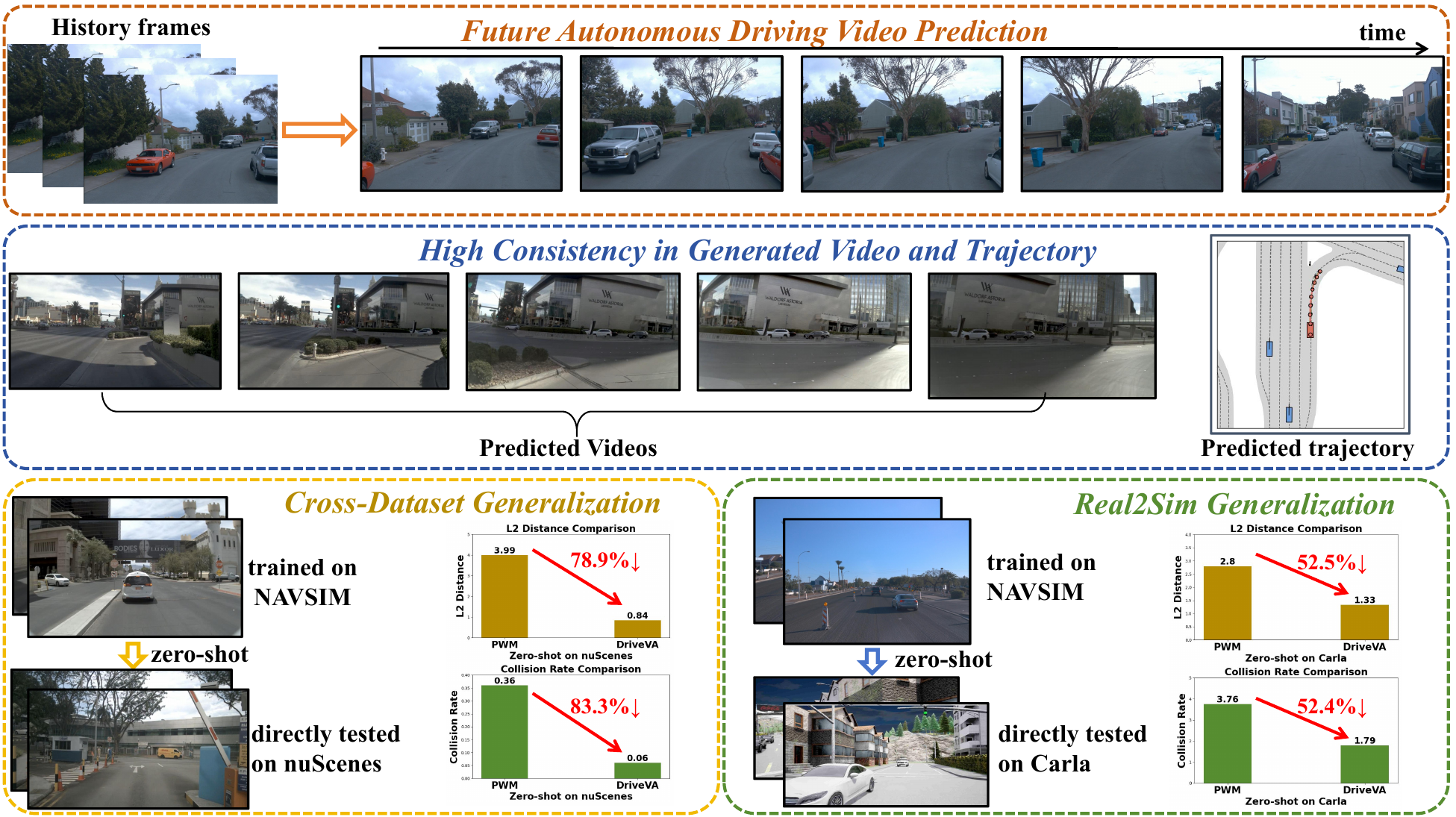}
    \vspace{-0.4cm}
\caption{\textbf{\ourmethod: unified video--trajectory rollout for planning.}
Given history frames, \ourmethod rolls out a future video clip (top).
The ego trajectory is generated together with the video rollout and remains aligned with the visual scene evolution (middle).
Bottom: zero-shot comparisons trained on NAVSIM and evaluated on nuScenes (cross dataset) and CARLA (cross domain from real to simulation), showing large relative improvements over PWM~\citep{zhao2025forecasting} in displacement error and collision rate.}
    \label{fig:first}
    \vspace{-0.3cm}
\end{figure}

\begin{abstract}
Generalization is a central challenge in autonomous driving, as real-world deployment requires robust performance under unseen scenarios, sensor domains, and environmental conditions. Recent world-model-based planning methods have shown strong capabilities in scene understanding and multi-modal future prediction, yet their generalization across datasets and sensor configurations remains limited. In addition, their loosely coupled planning paradigm often leads to poor video-trajectory consistency during visual imagination. To overcome these limitations, we propose \ourmethod, a novel autonomous driving world model that jointly decodes future visual forecasts and action sequences in a shared latent generative process. \ourmethod inherits rich priors on motion dynamics and physical plausibility from well-pretrained large-scale video generation models to capture continuous spatiotemporal evolution and causal interaction patterns. To this end, \ourmethod employs a DiT-based decoder to jointly predict future action sequences (trajectories) and videos, enabling tighter alignment between planning and scene evolution. We also introduce a video continuation strategy to strengthen long-duration rollout consistency. \ourmethod achieves an impressive PDM-based planning performance of 90.9 PDM score on the NAVSIM benchmark. Extensive experiments also demonstrate the zero-shot capability and cross-domain generalization of \ourmethod, which reduces average L2 error and collision rate by 78.9\% and 83.3\% on nuScenes and  52.5\% and 52.4\% on the Bench2Drive built on CARLA\,v2 compared with the state-of-the-art world-model-based planner.
\end{abstract}

\section{Introduction}
\label{sec:intro}

Generalization has long been a fundamental goal in autonomous driving, as it is essential for building systems that operate reliably in the real world~~\cite{hao2025driveaction,chi2026driver,liu2023regformer,du2026unsupervised,liu2024difflow3d,yu2026g2dp,cheng2023gatraj,liu2024laformer,liu2026streamvlo,feng2025survey_dwm_survey,xiong2026unidrive}. A capable autonomous driving model should not only perform well on scenarios seen during training, but also remain robust under unseen traffic patterns, novel road layouts, and diverse sensor configurations~\cite{li2023domain,zhou2026opendrivevla,hu2025vlm}. This ability is especially important for real-world deployment, where long-tail events, domain shifts, and complex agent interactions are common. Recent advances in large-scale pre-trained models have motivated researchers to develop autonomous driving systems that can better transfer across tasks and domains~\cite{yang2025drivemoe,liang2026render,liang2025multi}. This trend has led to the development of Vision-Language-Action (VLA) models~\cite{fu2025orion,fu2025minddrive,li2025drivevlaw0,yang2025drivemoe,zhou2026opendrivevla}, which leverage pre-trained vision-language models as the backbone and fine-tune them on driving-specific trajectory data. This strategy can reduce the amount of task-specific training data required while still achieving strong planning performance. However, despite these advances, true generalization, especially zero-shot transfer across datasets, remains limited and has yet to be fully realized. A key reason is that prevailing VLA pretraining on static image–text pairs primarily transfers semantic knowledge (“what is what”), but provides limited spatiotemporal and causal priors (“how the world moves”) needed for robust closed-loop planning.

Recently, large-scale video generation models~\cite{zheng2024opensora, yang2024cogvideox, kong2024hunyuanvideo, wan2025wan} have shown strong generalization to unseen textual prompts and visual contexts. By learning from massive video corpora, they capture realistic motion patterns and physically plausible scene dynamics~\cite{chen2026memobench,kong2024hunyuanvideo, wan2025wan,xiang2025macro,xiang2026pathwise,zhao2026luve,zhao2026geostream}, suggesting rich priors over real-world temporal evolution. 
Notably, the ability of video generators to produce temporally coherent future predictions under flexible conditioning aligns closely with the goal of building generalizable driving world models. This motivates a key question: \textbf{\emph{"Can large video generation models serve as a foundation for generalizable autonomous driving video action models?"}}

To answer this question, we investigate how to build and fine-tune autonom-ous driving models upon large-scale video generation models. 
Existing world-model-based planning methods suffer from two major bottlenecks. First, they often exhibit limited generalization across diverse datasets as the world knowledge learned from one dataset is difficult to transfer effectively to another.
Second, they commonly suffer from inconsistency between visual and action rollouts, because video imagination and trajectory generation are typically modeled separately or only loosely coupled~\cite{xia2025drivelaw,zhang2025epona}. 
To bridge the gap between generic video generation and driving-oriented planning, the key challenge is to enable video generation models not only to synthesize plausible future scenes, but also to produce actionable driving trajectories that can directly guide vehicle planning. Moreover, to effectively transfer the strong generalization capability of video generation models from the visual domain to the planning domain, it is crucial to maintain consistency between the predicted driving trajectories and the visual future evolution represented in the generated videos, as illustrated by the qualitative comparisons in Fig.~\ref{fig:visual_compare} and Fig.~\ref{fig:visual_dpvo_nuscenes}. Such alignment allows the semantic understanding and physical priors learned from large-scale video data to be naturally extended to autonomous driving behaviors.

In this paper, we propose a video-action model, called \ourmethod, which integrates large-scale video priors with end-to-end planning and dense supervision from world modeling. 
We find that video-level supervision is the main driver of planning gains, rather than a merely auxiliary loss appended to a pipeline as in most existing methods~\cite{xia2025drivelaw,zheng2025world4drive,zeng2025futuresightdrive,li2025drivevlaw0}. Concretely, enabling video supervision boosts NAVSIMv1 PDMS from 71.4 to 90.9 (+19.5) over action-only optimization (Table~\ref{tab:ablation}). The key is that video supervision provides dense temporal grounding of scene dynamics, and planning benefits only when the predicted actions are forced to stay consistent with the imagined future. 
As shown in Fig.~\ref{fig:first}, this motivates our unified formulation: instead of modeling future visual imagination and trajectory generation in separate stages, \ourmethod places future video latents and action tokens in a shared latent generative process and jointly decodes them with a single DiT~\citep{peebles2023DIT} in a shared latent space, so trajectories are generated as action grounding of the same rollout rather than being optimized in a separate stage. 
This unified formulation yields tighter video--trajectory alignment. 
Despite being generative, we observe that as few as two sampling steps already reach near-optimal planning performance, enabling efficient recurrent decision making. 
We further introduce a video continuation module to maintain long-duration consistency by progressively rolling out future video clips~\cite{liu2026arflow,liu20264dstr}. 
Extensive experiments demonstrate that \ourmethod achieves state-of-the-art PDM-based planning performance on NAVSIM, and also transfers strongly to unseen datasets across real driving scenes (e.g., nuScenes) and simulated scenes (e.g., Bench2Drive) in a zero-shot setting without target-domain fine-tuning, supporting the generalization capability of \ourmethod.

Overall, our core contributions are as follows:
\begin{itemize}
\item We propose \ourmethod, a unified video-action world model for autonomous driving that jointly models future visual imagination and trajectory prediction within a shared latent generative process, alleviating the mismatch caused by cascaded or loosely coupled planning pipelines.

\item We design a unified DiT-based decoder that simultaneously generates future video latents and action tokens, leading to stronger video-trajectory consistency and tighter alignment between scene evolution and planned behavior.

\item We introduce a video continuation module that progressively rolls out future clips, preserving long-horizon structural consistency during recurrent planning.

\item Extensive experiments show that \ourmethod achieves state-of-the-art PDM-based planning performance on NAVSIM (\textbf{90.9} PDMS), and delivers strong zero-shot performance on nuScenes (trained on NAVSIM) with \textbf{78.9\%} lower average L2 error and \textbf{83.3\%} lower collision rate than the state of the art. It also improves generalization from real to simulation on Bench2Drive (CARLA), reducing average L2 error by \textbf{52.5\%} and collision rate by \textbf{52.4\%}.
\end{itemize}

\section{Related Work}
\label{sec:related}

\subsection{Vision Language Action Models}
\textbf{VLAs.} Recently, the rapid development of vision-language-action (VLA) models~\cite{zhou2026opendrivevla,yang2025drivemoe,li2025drivevlaw0,fu2025minddrive,fu2025orion} has advanced a new paradigm for language-guided autonomous driving: these models jointly integrate language understanding, environment perception, and vehicle control to accomplish driving tasks in an end-to-end manner. This progress has been largely enabled by the continued evolution of vision~\cite{radford2021learning_clip, zhai2023sigmoid_siglip,oquab2023dinov2,xiao2025visual,xiao2026promptbased,guo2026vitexqa,du2026unsupervised}, language~\cite{team2024gemma,abdin2024phi3}, and vision-language~\cite{liu2023visual_llava, chen2024internvl, wang2024qwen2vl} foundation models. Despite this progress, most existing driving VLAs are still built upon vision-language models (VLMs) pre-trained on large-scale web data. While such models are effective at transferring visual-semantic knowledge, their pretraining data is primarily composed of \textit{static} image-text pairs, which limits their ability to capture temporal dynamics and physical interaction patterns directly; They do not naturally inherit the spatiotemporal priors required for adapting to new complex interactive scenarios. Consequently, the generalization capability of current driving VLAs, especially when faced with unseen scenarios and unseen behaviors, still exhibits clear limitations~\cite{zhou2026opendrivevla}.

\noindent\textbf{Generalization in VLAs.} To address the generalization issue, existing driving VLA methods mainly follow two directions: one focuses on targeted data construction for corner cases, while the other relies on structured expert modeling for long-tail behaviors~\cite{hao2025driveaction,hu2025vlm,zhou2026opendrivevla,chi2025impromptu}. However, stronger \emph{zero-shot} generalization remains insufficiently addressed. For example, Impromptu VLA~\cite{chi2025impromptu} improves robustness through a manually curated corner-case dataset~\cite{hu2025vlm,hao2025driveaction}, but relies on predefined scenario categories and trajectory-centric supervision, limiting true cross-dataset zero-shot transfer. DriveMoE~\cite{yang2025drivemoe}, in contrast, addresses rare and long-tail driving behaviors through scene- and skill-specialized experts~\cite{zhou2026opendrivevla}, yet still depends on predefined skill partitions and benchmark-specific data distributions, with limited evidence of transfer to unseen platforms or environments. We argue that \emph{zero-shot driving capability} is particularly critical for planning, as it more directly measures whether a model can make reliable decisions when encountering unseen corner cases rather than merely interpolating among observed trajectory patterns, and also serves as an indicator of cross-platform and cross-scenario generalization. In contrast, video-based world models can leverage dense frame-level supervision to learn physical dynamics from visual evolution, offering a more scalable path toward generalization beyond fixed action templates, benchmark-specific skill partitions, and manually defined corner-case taxonomies~\cite{wang2024driving, yang2024generalized, zheng2025world4drive, li2025drivevlaw0,gao2024vista}.

\subsection{Video Model-based Autonomous Driving}

Motivated by intuitive physical reasoning, world models aim to improve driving decisions by forecasting future scene evolution. Existing autonomous-driving world models can be broadly divided into two lines: latent-dynamics models for planning~\cite{zheng2025world4drive,yang2025raw2drive,wang2025adawm}, and models that explicitly predict future visual observations for decision making~\cite{zeng2025futuresightdrive,zhang2025epona,li2025drivevlaw0}.

 Latent-dynamics methods, such as LAW~\cite{li2024enhancing}, World4Drive~\cite{zheng2025world4drive}, AdaWM~\cite{wang2025adawm}, and Raw2Drive~\cite{yang2025raw2drive}, learn compact future representations for planning, policy optimization, or robustness, but generally treat the world model as an auxiliary module for supervision or planning guidance rather than using explicit visual rollouts at inference. In contrast, visually predictive approaches, including FutureSightDrive~\cite{zeng2025futuresightdrive}, Epona~\cite{zhang2025epona}, DriveVLA-W0~\cite{li2025drivevlaw0}, and DriveLaW~\cite{xia2025drivelaw}, more directly exploit future visual prediction for planning. However, existing methods still typically treat visual prediction as an auxiliary signal, an intermediate reasoning process, or a module loosely coupled with planning. Even in methods that connect the two more explicitly, video and action generation are often maintained as separate branches, with consistency relying on inter-module feature transfer or multi-stage optimization. As a result, mismatches between imagined futures and generated actions can accumulate over time, causing the executed actions to deviate from the future evolution predicted by the world model.

Built on video diffusion backbones, recent VAM-style driving models such as VaViM/VaVAM~\cite{bartoccioni2025vavim} explore video generative modeling for autonomous driving, suggesting that video generation priors can provide useful spatiotemporal and physical priors for driving policies. Unlike latent world models that learn dynamics from scratch in compact latent spaces~\citep{hafner2019dream, hafner2020mastering, hafner2023mastering, assran2025vjepa2}, they can directly exploit pretrained video representations that already encode rich physical dynamics. This suggests that a unified generative formulation of future video and actions may provide tighter coupling between visual forecasting and planning, while also improving transfer across data domains. Motivated by this observation, our method adopts a single shared generative process to jointly model future imagination and action generation, and further investigates how data scale and diversity affect generalization in autonomous driving.

\begin{figure}[t]
    \centering
    \includegraphics[width=1.0\linewidth]{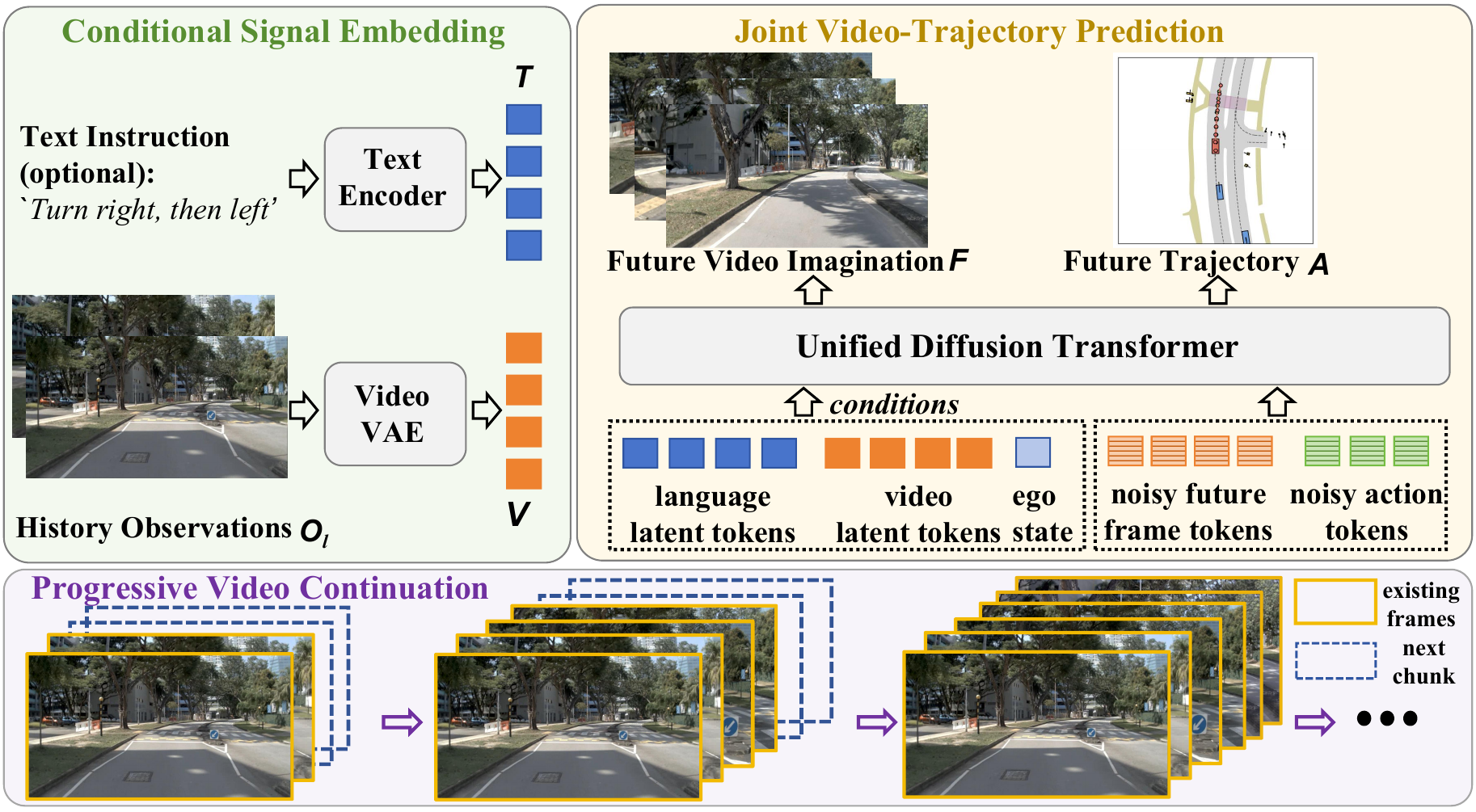}
    \caption{\textbf{Overall pipeline of \ourmethod.} 
    Given history observations, the ego state (current velocity $v_{x}, v_{y}$), and language instructions, the model first encodes conditional signals into latent tokens through a text encoder and a video VAE~\cite{wan2025wan}. 
    A unified diffusion transformer (DiT)~\cite{peebles2023DIT} then jointly predicts future video latents and future action tokens in a shared generative process, ensuring strong video–trajectory consistency. 
    To maintain long-horizon temporal coherence, a progressive video continuation strategy recursively rolls out future video clips while updating predicted trajectories.}
    \label{fig:pipeline}
    \vspace{-0.3cm}
\end{figure}

\section{Preliminary}
\label{sec:prelim}

\noindent
\textbf{Flow Matching.} Flow matching~\cite{lipman2022flow,liu2022flow,tong2024improving} models generation as a continuous-time transformation from a simple source distribution to the target data distribution. Let $x_{\mathrm{data}} \in \mathbb{R}^d$ be a data sample and $\boldsymbol{\epsilon} \sim \mathcal{N}(\mathbf{0},\mathbf{I})$ be a noise sample. The model learns a time-dependent velocity field $v_\theta:\mathbb{R}^d \times [0,1] \rightarrow \mathbb{R}^d$ that defines the dynamics of a trajectory $x^{(s)}$ via
\begin{equation}
\frac{d x^{(s)}}{d s} = v_\theta\!\left(x^{(s)}, s\right), \qquad x^{(0)}=\boldsymbol{\epsilon},\quad s\in[0,1].
\end{equation}
Intuitively, the learned flow transports samples from noise at $s=0$ to the data manifold at $s=1$.

For training, flow matching supervises the model on a prescribed interpolation path between $\boldsymbol{\epsilon}$ and $x_{\mathrm{data}}$. We use the standard linear interpolation $x^{(s)}=(1-s)\boldsymbol{\epsilon} + s x_{\mathrm{data}}$, whose derivative is $\dot{x}^{(s)} = x_{\mathrm{data}}-\boldsymbol{\epsilon}$. The network $v_\theta$ is trained to regress this target velocity:
\begin{equation}
\mathcal{L}_{\mathrm{FM}}
=
\mathbb{E}_{s,\boldsymbol{\epsilon},x_{\mathrm{data}}}
\left[
\left\|v_\theta\!\left(x^{(s)},s\right)-\dot{x}^{(s)}\right\|_2^2
\right].
\end{equation}

At inference, generation starts from Gaussian noise and integrates the learned velocity field from $s=0$ to $s=1$:
\begin{equation}
x^{(1)} = x^{(0)} + \int_0^1 v_\theta\!\left(x^{(s)},s\right)\, ds,
\qquad x^{(0)}\sim\mathcal{N}(\mathbf{0},\mathbf{I}).
\end{equation}

\noindent
\textbf{Video Generation with Conditional Flow Matching.}
Recent video generators~\cite{videoworldsimulators2024,wan2025wan} commonly perform flow matching in the latent space of a pretrained video autoencoder. Let $E$ and $D$ denote the encoder and decoder, respectively. Given a conditioning signal $c$, we aim to generate a latent video sequence $\mathbf{z}=\{z_1,\ldots,z_{T_v}\}$ and decode it to pixels with $D$.

Conditional flow matching learns a velocity field $v_\theta(\mathbf{z}^{(s)}, s \mid c)$ that defines the latent dynamics
$\frac{d\mathbf{z}^{(s)}}{ds}=v_\theta(\mathbf{z}^{(s)}, s \mid c)$ with $\mathbf{z}^{(0)}\sim\mathcal{N}(\mathbf{0},\mathbf{I})$.
Integrating from $s\!=\!0$ to $1$ yields the clean latent $\mathbf{z}^{(1)}$, which is then decoded by $D$.
This latent-space formulation is efficient and well-suited for long-horizon, condition-controlled video synthesis.

\section{Method}
\label{sec:method}

\subsection{Problem Formulation}

Given a language instruction $\mathcal{T}$ (including the high-level command) and a history observation buffer $\mathcal{O}_l=\{\mathbf{F}_{l-m+1},\ldots,\mathbf{F}_{l}\}$,
which contains $m$-frame history observations from $\mathbf{F}_{l-m+1}$ to $\mathbf{F}_l$. Our goal at the current timestep $l$ is to jointly predict future actions (trajectories) and future visual imaginations. Specifically, conditioned on $\mathcal{T}$, the current ego state $\mathbf{q}_l$ (represented by the ego velocity components $v_x$ and $v_y$), and the visual history observations $\mathcal{O}_l$, we predict:
\begin{enumerate}[leftmargin=1em]
    \item \textbf{An action chunk} $\mathcal{A}_{l+1:l+K}=\{\boldsymbol{a}_{l+i}\in\mathbb{R}^3\}_{i=1}^{K}$ consisting of $K$ future actions to be executed sequentially, where each action $\boldsymbol{a}_{l+i}$ is a 3-D vector. The first two dimensions encode the ego-vehicle $(x,y)$ position, and the last dimension encodes the yaw angle.
    \item \textbf{A future video clip} $\mathcal{F}_{l+1:l+N}=\{\boldsymbol{F}_{l+j}\}_{j=1}^{N}$ consisting of $N$ frames that depict the anticipated future visual evolution by executing $\mathcal{A}_{l+1:l+K}$. In practice, we do not predict raw frames directly; instead, we predict their latent representations, as detailed in Sec.~\ref{sec:preprocessing_updated}.
\end{enumerate}

After executing $\mathcal{A}_{l+1:l+K}$, we obtain new observations, update $\mathcal{O}_l$ using a sliding window, and repeat the process until task completion. This rolling-horizon setup reduces difficulty in long-horizon prediction to a progressive sequence of short \emph{video-continuation} problems.

\vspace{3pt}
\noindent\textbf{Joint Video--Action Modeling.}
Formally, \ourmethod jointly models future video imaginations and action chunks conditioned on $\mathbf{C}_l := (\mathcal{O}_l,\mathcal{T},\mathbf{q}_l)$.
This formulation can be viewed as unifying video continuation and IDM-style~\citep{du2023learning, zhou2024robodreamer} action grounding within a single end-to-end model, where actions are predicted to be consistent with the imagined future. Instead of training two separate models~\cite{pai2025mimic,lingbot-va2026,xia2025drivelaw} (a video prediction model and an inverse dynamics model) for the decomposed objective, we optimize a single model end-to-end with this joint objective. This design encourages tighter video--action alignment through deep cross-modal integration (Fig.~\ref{fig:visual1} and Fig.~\ref{fig:visual_dpvo_nuscenes}). Moreover, since pretrained video models already provide strong video-prediction priors from large web-scale data, \ourmethod focuses on adapting these priors to driving-domain video continuation and learning action grounding from predicted visual futures. We further hypothesize that this improves generalization power over conventional VLA training from VLMs, because our formulation explicitly learns temporal dynamics from video frames, which are both used as conditional inputs and prediction targets.

\subsection{Data Preprocessing}
\label{sec:preprocessing_updated}

\vspace{3pt}
\noindent\textbf{Text Instruction Encoding.}
We use a frozen text encoder from Wan2.2-TI2V-5B~\cite{wan2025wan} to encode the language instruction $\mathcal{T}$ (including the high-level command) into a fixed-length token sequence $
\mathbf{T}\in\mathbb{R}^{L_T\times d}$ (Fig.~\ref{fig:pipeline}). 
These encoded text tokens are injected into the backbone through a cross-attention mechanism, instead of being concatenated with the visual/action stream, keeping the spatiotemporal token sequence compact and decoupling text length for higher control flexibility.

\vspace{3pt}
\noindent\textbf{Video Causal VAE with Wan2.2-TI2V-5B.}
We adopt the 3D-causal VAE encoder from Wan2.2-TI2V-5B as the video encoder. Given a video clip
$\mathcal{F}=\{\mathbf{F}_j\}_{j=1}^{N}$, the encoder produces a temporally downsampled latent sequence:
\begin{equation}
\mathcal{V}=\{\mathbf{V}_j\}_{j=1}^{n}, \qquad \mathbf{V}_j\in\mathbb{R}^{h\times w\times c},
\end{equation}
where $n$ is the latent sequence length after temporal downsampling.
In the original WAN's design, causality ensures that the first latent feature $\mathbf{V}_1$ depends only on the first frame observation $\mathbf{F}_1$, so a single observed frame can be encoded as a valid conditional latent at inference time.

To guarantee long-duration consistency, we further extend this single-frame conditioning to a \emph{video-continuation} setting by conditioning on a history observation buffer rather than only the current frame. Specifically, at current timestep $l$, we encode the observation buffer
$\mathcal{O}_l=\{\mathbf{F}_{l-m+1},\ldots,\mathbf{F}_l\}$ into a sequence of history latents: $\mathcal{V}^{\mathrm{his}}_l=\{\mathbf{V}_{l-m+1},\ldots,\mathbf{V}_{l}\}.$
Thus, the first $m$-frame conditioning latents are all derived from historical observations and provide long-range visual priors for continuation. During training, we encode the full clip to obtain both history latents $\mathcal{V}^{\mathrm{his}}_l$ and future latents $\mathcal{V}^{\mathrm{fut}}_l$; during inference, we encode only $\mathcal{O}_l$ and generate future latents conditioned on the encoded history latents $\mathcal{V}^{\mathrm{his}}_l$.

\subsection{Consistent Video-Action Generation}
\label{sec:unified_future_updated}

At each current timestep $l$, \ourmethod  jointly predicts future video latents and action tokens conditioned on
(i) history latents $\mathcal{V}^{\mathrm{his}}_l$,
(ii) the current ego state $\mathbf{q}_l$, and
(iii) text/command tokens $\mathbf{T}$, as in Fig.~\ref{fig:pipeline}.
This design matches training and inference: both use a fixed-length history buffer and predict a short continuation window, which can be chained to progressively generate long-horizon rollout.

\vspace{3pt}
\noindent\textbf{Input Tokenization.}
We raster-flatten each visual latent $\mathbf{V}_t$ in $\mathcal{V}^{\mathrm{his}}$ and project it into the model dimension:
\begin{equation}
\mathbf{V}'_t=\mathrm{Proj}\!\left(\mathrm{Flatten}(\mathbf{V}_t)\right)\in\mathbb{R}^{L_V\times d},
\qquad L_V=h\cdot w.
\end{equation}
The current ego state $\mathbf{q}_l$ is embedded into $L_S$ state tokens
$\mathbf{S}_l\in\mathbb{R}^{L_S\times d}$ using an MLP.
Each action $\mathbf{a}_{l+i}\in\mathbb{R}^{3}$ is also embedded into one token via an MLP, yielding the action-token sequence $\mathbf{A}_{l+1:l+K}\in\mathbb{R}^{K\times d}$.

\vspace{3pt}
\noindent\textbf{Fixed Condition and Generative Targets.}
We split the model input into a \emph{noise-free condition block} and a \emph{generative target block}:
\begin{align}
\mathbf{X}_{\mathrm{cond}}^{(l)}
&=
[\,\mathbf{S}_l,\ \mathbf{V}'_{l-m+1},\ldots,\mathbf{V}'_{l}\,]
\in\mathbb{R}^{L_{\mathrm{cond}}\times d}, \\
\mathbf{Y}_{0}^{(l)}
&=
[\,\mathbf{V}'_{l+1},\ldots,\mathbf{V}'_{l+n_{\mathrm{pred}}},\ \mathbf{A}_{l+1:l+K}\,]
\in\mathbb{R}^{L_{\mathrm{tgt}}\times d}.
\end{align}

Here, $n_{\mathrm{pred}}$ is the number of future latent steps corresponding to the predicted future clip (after temporal downsampling). 
The condition block $\mathbf{X}_{\mathrm{cond}}^{(l)}$ is kept fixed at both training and inference. 
Given the conditional tokens $\mathbf{X}_{\mathrm{cond}}^{(l)}$ and text tokens $\mathbf{T}$, a Diffusion Transformer (DiT) decoder predicts the conditional velocity field for the generative targets:
\begin{equation}
\hat{\mathbf{v}}_{\theta}^{(l,s)}
=
f_{\theta}\!\left([\mathbf{X}_{\mathrm{cond}}^{(l)},\mathbf{Y}^{(l,s)}],\, s \mid \mathbf{T}\right),
\end{equation}
where $\mathbf{Y}^{(l,s)}$ is the noisy interpolation of the clean targets $\mathbf{Y}_0^{(l)}$ at flow time $s$, and $f_{\theta}$ is the DiT decoder parameterized by $\theta$.

\subsection{Flow Matching Objective}
\label{sec:flow_matching}

Following the flow matching formulation in Sec.~\ref{sec:prelim}, we instantiate a \emph{conditional} flow over the generative target block $\mathbf{Y}_0^{(l)}$, conditioned on the fixed context block $\mathbf{X}_{\mathrm{cond}}^{(l)}$ and text tokens $\mathbf{T}$.

\vspace{3pt}
\noindent\textbf{Flow-Matching Generative Loss.}
At timestep $l$, we denote the clean target tokens as $\mathbf{Y}_0^{(l)}$. 
We sample $s\sim\mathcal{U}(0,1)$ and $\boldsymbol{\epsilon}\sim\mathcal{N}(\mathbf{0},\mathbf{I})$, and construct the linear interpolation
$\mathbf{Y}^{(l,s)}=(1-s)\boldsymbol{\epsilon} + s\mathbf{Y}_0^{(l)},$
whose target velocity is
$\dot{\mathbf{Y}}^{(l,s)}=\mathbf{Y}_0^{(l)}-\boldsymbol{\epsilon}.$
We optimize the standard flow-matching regression loss:
\begin{equation}
\mathcal{L}_{\mathrm{FM}}
=
\mathbb{E}_{l,s,\mathbf{Y}_0^{(l)},\boldsymbol{\epsilon}}
\left[
\left\|\hat{\mathbf{v}}_{\theta}^{(l,s)}-\dot{\mathbf{Y}}^{(l,s)}\right\|_2^2
\right].
\end{equation}

\section{Experiments}

\subsection{Datasets}

\noindent\textbf{NAVSIM v1.}
We use the NAVSIM\,v1 benchmark~\citep{dauner2024navsim} (built on OpenScene~\citep{contributors2023openscene}) as our main PDM-based evaluation for safety-critical driving.
It reports NC, DAC, TTC, Comfort (C.), and Ego Progress (EP), aggregated as
\( \mathrm{PDMS} = \mathrm{NC}\times \mathrm{DAC}\times \frac{5\mathrm{EP}+5\mathrm{TTC}+2\mathrm{C.}}{12} \).

\noindent\textbf{nuScenes.} For cross-dataset zero-shot evaluation, we evaluate on the nuScenes validation split (150 scenes) from the 1,000-scene nuScenes dataset~\citep{caesar2020nuscenes}, and report Displacement Error (DE) and Collision Rate (CR) following prior works~\cite{hu2023planning,jiang2023vad}. 

\noindent\textbf{Bench2Drive.} Bench2Drive~\citep{jia2024bench2drive} is a CARLA\,v2 closed-loop benchmark~\citep{dosovitskiy2017carla} with diverse interactive scenarios and evaluation routes. We evaluate: (1) From real to simulation cross domain zero-shot transfer by testing a NAVSIM-trained model directly on the Bench2Drive validation split; (2) \emph{sim-enhanced} training by mixing NAVSIM and Bench2Drive data, then evaluating on NAVSIM. Note that transferring policies across real-world logs and simulation is challenging due to the well-known \emph{reality gap} in appearance, dynamics, and agent behaviors~\citep{hu2023simulation}.

\begin{table}[t!]
    \centering
    \caption{\textbf{Performance comparison on NAVSIM \textit{Navtest} using PDM-based metrics.}
    Methods are grouped by whether they employ an explicit world model: \textit{Traditional End-to-End Methods} and \textit{World Model Methods}.}
    \setlength{\tabcolsep}{6pt}
    \label{tab:navsim1}
    \vspace{-0.2cm}
    \resizebox{0.98\textwidth}{!}{%
    \begin{tabular}{@{}l|c|cc|cc|cccc@{}}
        \toprule
        \textbf{Method} & \textbf{Ref} & \textbf{Image} & \textbf{Lidar} &
        \textbf{NC$\uparrow$} & \textbf{DAC$\uparrow$} &
        \textbf{TTC$\uparrow$} & \textbf{Comf.$\uparrow$} & \textbf{EP$\uparrow$} & \textbf{PDMS$\uparrow$} \\
        \midrule
        Constant Velocity & - &  &  & 68.0 & 57.8 & 50.0 & \textbf{100} & 19.4 & \cellcolor{gray!30}20.6 \\
        Ego Status MLP\cite{dauner2024navsim} & arXiv'23 &  &  & 93.0 & 77.3 & 83.6 & \textbf{100} & 62.8 & \cellcolor{gray!30}65.6 \\
        \midrule

        \rowcolor{gray!30}\multicolumn{10}{@{}l}{\raggedright \textit{Traditional End-to-End Methods}} \\
        VADv2-$\mathcal{V}_{\text{8192}}$~\citep{jiang2026vadv} & ICLR'26 & \cmark &  & 97.2 & 89.1 & 91.6 & \textbf{100} & 76.0 & \cellcolor{gray!30}80.9 \\
        UniAD~\citep{hu2023planning} & CVPR'23 & \cmark &  & 97.8 & 91.9 & 92.9 & \textbf{100} & 78.8 & \cellcolor{gray!30}83.4 \\
        TransFuser~\citep{chitta2022transfuser} & TPAMI'23 & \cmark & \cmark & 97.7 & 92.8 & 92.8 & \textbf{100} & 79.2 & \cellcolor{gray!30}84.0 \\
        PARA-Drive~\citep{weng2024drive} & CVPR'24 & \cmark &  & 97.9 & 92.4 & 93.0 & 99.8 & 79.3 & \cellcolor{gray!30}84.0 \\
        ReCogDrive-IL~\citep{li2025recogdrive} & ICLR'26 & \cmark &  & 98.1 & 94.7 & 94.2 & \textbf{100} & 80.9 & \cellcolor{gray!30}86.5 \\
        DiffusionDrive~\citep{liao2025diffusiondrive} & CVPR'25 & \cmark & \cmark & 98.2 & 96.2 & 94.7 & \textbf{100} & 82.2 & \cellcolor{gray!30}88.1 \\
        \midrule

        \rowcolor{gray!30} \multicolumn{10}{@{}l}{\raggedright \textit{World Model Methods}} \\
        DrivingGPT~\citep{chen2025drivinggpt} & ICCV'25 & \cmark &  & 98.9 & 90.7 & 94.9 & 95.6 & 79.7 & \cellcolor{gray!30}82.4 \\
        LAW~\citep{li2024enhancing} & ICLR'25 & \cmark &  & 96.4 & 95.4 & 88.7 & 99.9 & 81.7 & \cellcolor{gray!30}84.6 \\
        Epona~\citep{zhang2025epona} & ICCV'25 & \cmark &  & 97.9 & 95.1 & 93.8 & 99.9 & 80.4 & \cellcolor{gray!30}86.2 \\
        Resim~\citep{yang2025resim} & NeurIPS'25 & \cmark &  & -- & -- & -- & -- & -- & \cellcolor{gray!30}86.6 \\
        WoTE~\citep{li2025end_wote} & ICCV'25 & \cmark & \cmark & 98.5 & 96.8 & 94.9 & 99.9 & 81.9 & \cellcolor{gray!30}88.3 \\
        DriveVLA-W0~\citep{li2025drivevlaw0} & ICLR'26 & \cmark &  & 98.4 & 95.3 & 95.2 & \textbf{100} & 80.9 & \cellcolor{gray!30}87.2 \\
        PWM~\citep{zhao2025forecasting} & NeurIPS'25 & \cmark &  & 98.6 & 95.9 & 95.4 & \textbf{100} & 81.8 & \cellcolor{gray!30}88.1 \\
    \textbf{Ours} & - & \cmark &  & \textbf{99.2} & \textbf{97.5} & \textbf{98.7} & \textbf{100} & \textbf{83.5} & \cellcolor{gray!30}\textbf{90.9} \\
        \bottomrule
    \end{tabular}%
    }
    \vspace{-0.3cm}
    \label{tab:comparison_modified}
\end{table}

\subsection{Training Details} 
We utilize Wan2.2-TI2V-5B~\cite{wan2025wan} as our pre-trained backbone. Each training sample consists of $4$ history frames and $8$ future frames at $2$ FPS with a resolution of $832\times480$. Training is performed on NVIDIA H20 GPUs with AdamW using a learning rate of $10^{-4}$ and weight decay of $0.01$ under distributed bf16 mixed-precision training. We first train with a batch size of $80$ for $20$k steps for faster convergence, and then continue fine-tuning for another $10$k steps with an effective batch size of $640$ via gradient accumulation. We adopt a linear warm-up over the first $1k$ steps starting from $10^{-3}$ of the base learning rate, followed by a constant learning rate schedule. The training objective combines a flow-matching loss for future-frame generation with a trajectory prediction loss. During inference, we use 2 sampling steps for flow-based sampling.

\subsection{Quantitative Comparison Results}

\noindent\textbf{Comparison Results on NAVSIM v1.} As shown in Table \ref{tab:navsim1}, we evaluate our \ourmethod on the NAVSIM dataset using PDM-based metrics. As for the average PDMS metric, our method not only surpasses previous traditional end-to-end methods like DiffusionDrive~\cite{liao2025diffusiondrive}, but also outperforms recent world model-based methods, including latent-dynamics approaches such as LAW~\cite{li2024enhancing} and visually predictive models that explicitly forecast future observations for decision making, e.g., Epona~\cite{zhang2025epona}, PWM~\cite{zhao2025forecasting} and DriveVLA-W0~\cite{li2025drivevlaw0}, etc. Notably, compared to some recent methods such as WoTE \cite{li2025end_wote} using multi-modal information as inputs, our method uses only front-view camera images, but still achieves a better score on the PDM-based planning benchmark. We attribute these gains primarily to our unified video-action formulation, which jointly models future video imagination and ego trajectory prediction within a shared generative process, leading to better alignment between what the model imagines and how it plans.

\begin{table*}[t]
\setlength{\tabcolsep}{1.5pt}
\caption{\textbf{Zero-shot end-to-end motion planning performance on nuScenes~\cite{caesar2020nuscenes} and Bench2Drive (CARLA)~\cite{jia2024bench2drive}}. 
All methods are trained on NAVSIM and directly evaluated on the target datasets without target-domain fine-tuning.}
\vspace{-0.2cm}
\centering
\resizebox{0.98\linewidth}{!}{
\begin{tabular}{l|cc|cccc|cccc|cccc|cccc}
\toprule
\multirow{2}{*}{Method} &
\multirow{2}{*}{Finetune} &
\multirow{2}{*}{Ref} &
\multicolumn{8}{c|}{nuScenes} &
\multicolumn{8}{c}{Bench2Drive (CARLA)} \\

\cline{4-19}

&&&
\multicolumn{4}{c|}{L2 (m) $\downarrow$} &
\multicolumn{4}{c|}{Collision (\%) $\downarrow$} &
\multicolumn{4}{c|}{L2 (m) $\downarrow$} &
\multicolumn{4}{c}{Collision (\%) $\downarrow$} \\

&&&
1s & 2s & 3s & \cellcolor{gray!30}Avg. &
1s & 2s & 3s & \cellcolor{gray!30}Avg. &
1s & 2s & 3s & \cellcolor{gray!30}Avg. &
1s & 2s & 3s & \cellcolor{gray!30}Avg. \\

\midrule
\rowcolor{gray!30}\multicolumn{2}{@{}l}{\raggedright \textit{VLA-World Model Methods}} \\
DriveVLA-W0~\citep{li2025drivevlaw0}
& \xmark
& ICLR'26
& 0.43 & 1.26 & 2.60 & \cellcolor{gray!30}1.43
& 0.22 & 0.66 & 1.42 & \cellcolor{gray!30}0.77
& 1.01 & 2.77 & 5.22 & \cellcolor{gray!30}3.00
& 1.49 & 2.53 & 3.53 & \cellcolor{gray!30}2.52
\\

\rowcolor{gray!30}\multicolumn{2}{@{}l}{\raggedright \textit{World Model Methods}} \\

PWM~\citep{zhao2025forecasting}
& \xmark
& NeurIPS'25
& 2.06 & 3.91 & 6.00 & \cellcolor{gray!30}3.99
& 0.12 & 0.15 & 0.86 & \cellcolor{gray!30}0.36
& 1.70 & 2.74 & 3.97 & \cellcolor{gray!30}2.80
& 4.01 & 3.73 & 3.53 & \cellcolor{gray!30}3.76
\\

\textbf{Ours}
& \textbf{\xmark}
& -
& \textbf{0.33} & \textbf{0.76} & \textbf{1.43} & \cellcolor{gray!30}\textbf{0.84}
& \textbf{0.00} & \textbf{0.07} & \textbf{0.12} & \cellcolor{gray!30}\textbf{0.06}
& \textbf{0.69} & \textbf{1.29} & \textbf{2.03} & \cellcolor{gray!30}\textbf{1.33}
& \textbf{1.38} & \textbf{1.97} & \textbf{2.65} & \cellcolor{gray!30}\textbf{1.79}
\\

\bottomrule
\end{tabular}}
\label{tab:zero_shot}
\end{table*}

\begin{table*}[t]
\setlength{\tabcolsep}{0.01\linewidth}
\caption{\textbf{End-to-end motion planning performance on nuScenes~\cite{caesar2020nuscenes} dataset.} 
$^*$ represents only using the front camera as input.}
\vspace{-0.2cm}
\centering
\resizebox{0.98\linewidth}{!}{
\begin{tabular}{l|cc|lc|cccc|cccc}
\toprule
\multirow{2}{*}{Method} & \multirow{2}{*}{\shortstack{nuScenes\\Finetune}} & \multirow{2}{*}{Ref} & \multirow{2}{*}{Input} & \multirow{2}{*}{Auxiliary Supervision} &
\multicolumn{4}{c|}{L2 (m) $\downarrow$} & 
\multicolumn{4}{c}{Collision Rate (\%) $\downarrow$} \\
&&&&& 1s & 2s & 3s & \cellcolor{gray!30}Avg. & 1s & 2s & 3s & \cellcolor{gray!30}Avg.  \\
\midrule
ST-P3~\cite{hu2022stp3} & \cmark & ECCV'22 & Camera & Map\&Box\&Depth & 1.33 & 2.11 & 2.90 & \cellcolor{gray!30}2.11 & 0.23 & 0.62 & 1.27 & \cellcolor{gray!30}0.71  \\
UniAD~\cite{hu2023planning} & \cmark & CVPR'23 & Camera & {\footnotesize Map\&Box\&Motion} & {0.48} & {0.96} & {1.65} & \cellcolor{gray!30}{1.03} & {0.05} & 0.17 & 0.71 & \cellcolor{gray!30}{0.31}  \\
OccNet~\cite{tong2023scene} & \cmark & ICCV'23 & Camera & 3D-Occ\&Map\&Box & 1.29 & 2.13 & 2.99 & \cellcolor{gray!30}2.14 & 0.21 & 0.59 & 1.37 & \cellcolor{gray!30}0.72  \\
OccWorld~\cite{zheng2024occworld} & \cmark & ECCV'24 & Camera & 3D-Occ & 0.52 & 1.27 & 2.41 & \cellcolor{gray!30}1.40 & 0.12 & 0.40 & 2.08 & \cellcolor{gray!30}0.87  \\
VAD-Tiny~\cite{jiang2023vad} & \cmark & ICCV'23 & Camera & Map\&Box\&Motion & 0.60 & 1.23 & 2.06 & \cellcolor{gray!30}1.30 & 0.31 & 0.53 & 1.33 & \cellcolor{gray!30}0.72  \\
VAD-Base~\cite{jiang2023vad} & \cmark & ICCV'23 & Camera & Map\&Box\&Motion & 0.54 & 1.15 & 1.98 & \cellcolor{gray!30}1.22 & 0.04 & 0.39 & 1.17 & \cellcolor{gray!30}0.53 \\
GenAD~\cite{zheng2024genad} & \cmark & ECCV'24 & Camera & Map\&Box\&Motion & {0.36} & {0.83} & {1.55} & \cellcolor{gray!30}{0.91} & 0.06 & {0.23} & {1.00} & \cellcolor{gray!30}{0.43} \\
\midrule
Doe-1~\cite{zheng2024doe} & \cmark & arXiv'24 & Camera$^*$ & QA & 0.50 & 1.18 & 2.11 & \cellcolor{gray!30}1.26 & 0.04 & 0.37 & 1.19 & \cellcolor{gray!30}0.53  \\
Epona~\cite{zhang2025epona} & \cmark & ICCV'25 & Camera$^*$ & None & 0.61 & 1.17 & 1.98 & \cellcolor{gray!30}1.25 & 0.01 & 0.22 & 0.85 & \cellcolor{gray!30}0.36  \\
\textbf{Ours} & \textbf{\xmark} & - & Camera$^*$ & None & \textbf{0.33} & \textbf{0.76} & \textbf{1.43} & \cellcolor{gray!30}\textbf{0.84} & \textbf{0.00} & \textbf{0.07} & \textbf{0.12} & \cellcolor{gray!30}\textbf{0.06}  \\
\bottomrule
\end{tabular}%
}
\label{tab:plan_nusc_main}
\vspace{-0.3cm}
\end{table*}

\subsection{Cross-Domain Generalization}

\noindent \textbf{Zero-shot Evaluation}.
Table~\ref{tab:zero_shot} evaluates strict zero-shot transfer, where all methods are trained on NAVSIM and directly evaluated on nuScenes and Bench2-Drive without fine-tuning. On nuScenes, \ourmethod achieves the best performance among world-model methods across all horizons, with an average L2 error of 0.84 and an average collision rate of 0.06. Under the same protocol, camera-only input, and without auxiliary supervision, it substantially improves over PWM~\citep{zhao2025forecasting}, reducing the average L2 error from 3.99 to 0.84 (78.9\%) and the average collision rate from 0.36 to 0.06 (83.3\%).

On Bench2Drive, \ourmethod also shows strong robustness under real-to--simulation transfer. Compared with PWM, it lowers the average L2 error from 2.80 to 1.33 and the average collision rate from 3.76 to 1.79. It further outperforms DriveVLA-W0~\citep{li2025drivevlaw0}, reducing the average L2 error from 3.00 to 1.33 (55.7\%) and the average collision rate from 2.52 to 1.79 (29.0\%). Moreover, Table~\ref{tab:plan_nusc_main} shows that \ourmethod, even without nuScenes training or fine-tuning, surpasses baselines trained or fine-tuned on nuScenes, indicating that unified video-action modeling learns transferable planning priors rather than relying on target-domain adaptation.

\begin{figure}[t]
    \centering
    \includegraphics[width=1.0\linewidth]{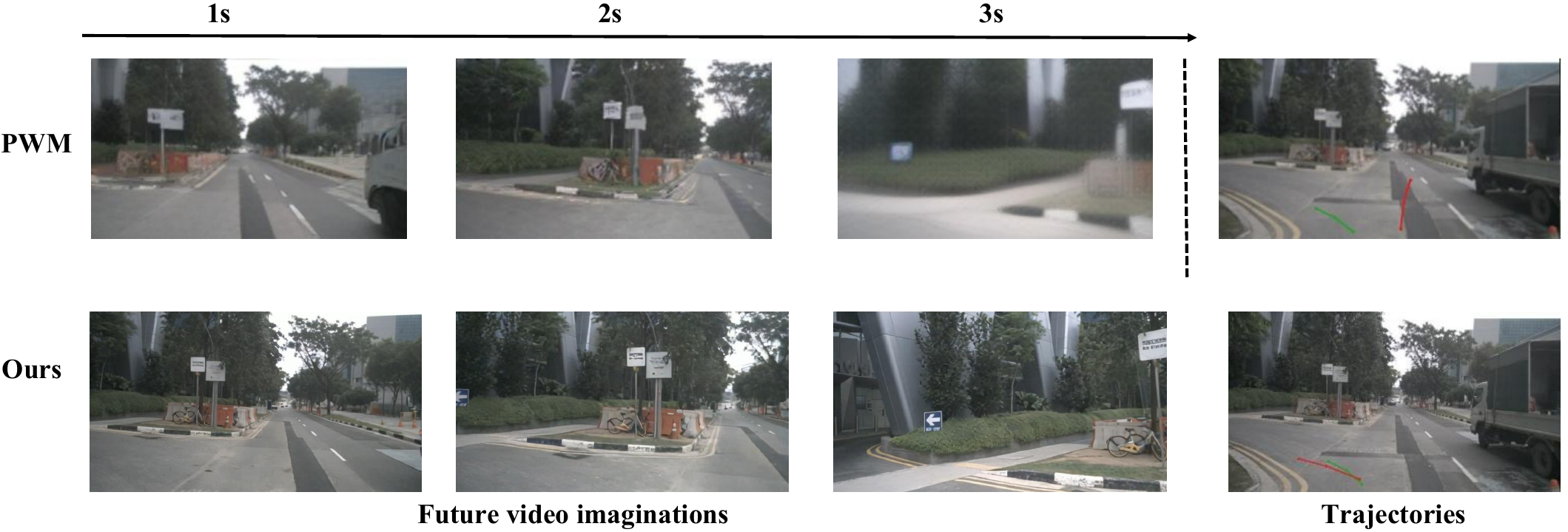}
\caption{\textbf{Video--trajectory consistency comparison in zero-shot unseen nuScenes scenarios.} 
In this left-turn scenario, our method produces trajectories that follow the scene evolution in the generated future video. 
In contrast, PWM~\cite{zhao2025forecasting} predicts a straight trajectory while the generated video indicates a left-turn maneuver, revealing a clear video–trajectory mismatch. Here, \textcolor{green}{green} denotes the GT trajectory and \textcolor{red}{red} the predicted trajectory.}
    \label{fig:visual_compare}
    \vspace{-0.1cm}
\end{figure}

\subsection{Video--Trajectory Consistency Analysis}

\begin{figure}[t]
    \centering
    \includegraphics[width=0.98\linewidth]{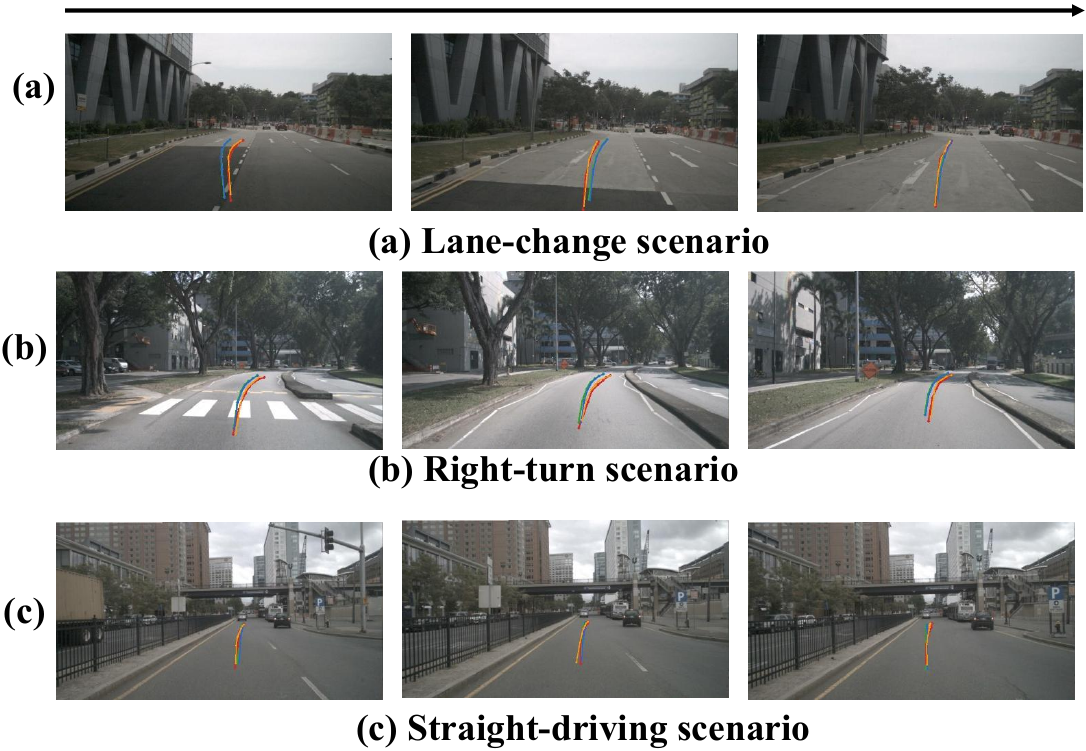}
    \caption{\textbf{DPVO-based qualitative analysis of video--trajectory consistency on nuScenes.}
    We visualize zero-shot nuScenes scenarios with temporal frames, including lane-change, right-turn, and straight-driving cases.
    \textcolor{green}{GT Future} and \textcolor{red}{Pred Future} denote the ground-truth and predicted trajectories, while \textcolor{dpvogat}{DPVO(gt img)} and \textcolor{dpvopred}{DPVO(pred img)} denote DPVO reconstructions from the ground-truth and predicted future videos.
    The close alignment among these curves provides qualitative evidence of video--trajectory consistency in \ourmethod.}
    \label{fig:visual_dpvo_nuscenes}
    \vspace{-0.55cm}
\end{figure}

To verify video--trajectory consistency, we reconstruct ego motion from ground-truth and generated future videos using DPVO~\cite{teed2023deep}. After aligning each reconstructed trajectory to its reference with a 2D similarity transform, we compute the average L2 error over the future 4s horizon. Table~\ref{tab:dpvo_consistency} shows consistently low errors on NAVSIM and zero-shot nuScenes, indicating that generated videos imply motion aligned with the predicted trajectories. Fig.~\ref{fig:visual_dpvo_nuscenes} further shows that DPVO reconstructions from generated videos closely follow model predictions across representative scenarios. More details and analyses are provided in Sec.~S4 of the supplementary material.

\begin{table}[t]
    \centering
    \caption{\textbf{Quantitative video--trajectory consistency measured by DPVO reconstruction.}
    We run DPVO~\citep{teed2023deep} on ground-truth future videos and generated future videos to reconstruct camera trajectories.
    After 2D similarity alignment, we compute the average L2 error over the future 4s horizon with respect to the corresponding reference trajectories.
    Lower is better.}
    \label{tab:dpvo_consistency}
    \vspace{-0.2cm}
    \setlength{\tabcolsep}{5pt}
    \renewcommand{\arraystretch}{1.05}
    \resizebox{0.95\columnwidth}{!}{
    \begin{tabular}{l|cc}
        \toprule
        Split / Scenario & GT traj. vs.\ GT-video recon. Avg. L2 (4s) $\downarrow$ & Pred. traj. vs.\ Pred.-video recon. Avg. L2 (4s) $\downarrow$ \\
        \midrule
        NAVSIM & 0.09 & 0.16 \\
        nuScenes & 0.07 & 0.14 \\
        \midrule
        Average & 0.08 & 0.15 \\
        \bottomrule
    \end{tabular}}
    \vspace{-0.2cm}
\end{table}

\begin{table*}[t]
\centering
\caption{\textbf{Ablation studies.} 
"Video Loss" indicates video supervision during training. 
"CARLA Mix Training" denotes joint training with CARLA simulated data and NAVSIM data. 
"Video Continuation" refers to video-to-video generation.}
\label{tab:ablation}

\vspace{-0.3cm}

\renewcommand\tabcolsep{3.9pt}
\resizebox{0.98\textwidth}{!}{
\begin{tabular}{c|ccc|cccccc}
\toprule
\multirow{2}{*}{ID} & \multirow{2}{*}{Video Loss} & \multirow{2}{*}{CARLA Mix Training} & \multirow{2}{*}{Video Continuation} & \multicolumn{6}{c}{Planning Metric} \\
 & & & & NC$\uparrow$ & DAC$\uparrow$ & TTC$\uparrow$ & Comf.$\uparrow$ & EP$\uparrow$ & \cellcolor{gray!30}PDMS$\uparrow$ \\
\midrule
1 & \xmark & \cmark & \cmark & 95.0 & 89.0 & 93.9 & 86.6 & 59.7 & \cellcolor{gray!30}71.4 \\
2 & \cmark & \xmark & \cmark & 99.0 & 97.3 & 98.4 & 100 & 83.2 & \cellcolor{gray!30}90.5 \\
3 & \cmark & \cmark & \xmark & 94.9 & 95.6 & 94.2 & 100 & 76.9 & \cellcolor{gray!30}84.6 \\
4 & \cmark & \cmark & \cmark & \textbf{99.2} & \textbf{97.5} & \textbf{98.7} & \textbf{100} & \textbf{83.5} & \cellcolor{gray!30}\textbf{90.9} \\
\bottomrule
\end{tabular}
}

\vspace{3mm}


\begin{minipage}[t]{0.485\textwidth}
\centering
\captionof{table}{\textbf{Future video frames.}}
\label{tab:futureframes}
\vspace{-0.2cm}
\scriptsize
\resizebox{\textwidth}{!}{
\begin{tabular}{p{2.2cm}|cccccc}
\toprule
Future Frames & NC$\uparrow$ & DAC$\uparrow$ & TTC$\uparrow$ & Comf.$\uparrow$ & EP$\uparrow$ & \cellcolor{gray!30}PDMS$\uparrow$ \\
\midrule
4  & 96.6 & 91.4 & 95.5 & 93.3 & 77.2 & \cellcolor{gray!30}82.1 \\
8  & \textbf{99.2} & \textbf{97.5} & \textbf{98.7} & \textbf{100} & \textbf{83.5} & \cellcolor{gray!30}\textbf{90.9} \\
12 & 98.6 & 94.4 & 97.5 & 99.8 & 79.5 & \cellcolor{gray!30}86.7 \\
\bottomrule
\end{tabular}
}
\end{minipage}
\hfill
\begin{minipage}[t]{0.485\textwidth}
\centering
\captionof{table}{\textbf{Training strategy.}}
\label{tab:trainstrategy}
\vspace{-0.2cm}
\scriptsize
\resizebox{\textwidth}{!}{
\begin{tabular}{p{2.2cm}|cccccc}
\toprule
Training Strategy & NC$\uparrow$ & DAC$\uparrow$ & TTC$\uparrow$ & Comf.$\uparrow$ & EP$\uparrow$ & \cellcolor{gray!30}PDMS$\uparrow$ \\
\midrule
From Scratch   & 89.9 & 76.8 & 87.6 & 99.9 & 76.8 & \cellcolor{gray!30}62.9 \\
LoRA Fine-tune & 92.4 & 88.0 & 91.0 & 99.9 & 67.5 & \cellcolor{gray!30}74.9 \\
Full Fine-tune & \textbf{99.2} & \textbf{97.5} & \textbf{98.7} & \textbf{100} & \textbf{83.5} & \cellcolor{gray!30}\textbf{90.9} \\
\bottomrule
\end{tabular}
}
\end{minipage}

\par\vspace{8pt}

\begin{minipage}[t]{0.485\textwidth}
\centering
\captionof{table}{\textbf{Sampling steps.}}
\label{tab:step}
\vspace{-0.2cm}
\scriptsize
\resizebox{\textwidth}{!}{
\begin{tabular}{p{2.2cm}|cccccc}
\toprule
Steps & NC$\uparrow$ & DAC$\uparrow$ & TTC$\uparrow$ & Comf.$\uparrow$ & EP$\uparrow$ & \cellcolor{gray!30}PDMS$\uparrow$ \\
\midrule
1 & 61.8 & 36.9 & 50.3 & 1.9 & 36.9 & \cellcolor{gray!30}13.2 \\
2 & \textbf{99.2} & \textbf{97.5} & \textbf{98.7} & \textbf{100} & 83.5 & \cellcolor{gray!30}\textbf{90.9} \\
3 & 99.1 & 97.4 & \textbf{98.7} & \textbf{100} & \textbf{83.7} & \cellcolor{gray!30}\textbf{90.9} \\
\bottomrule
\end{tabular}
}
\end{minipage}
\hfill
\begin{minipage}[t]{0.485\textwidth}
\centering
\captionof{table}{\textbf{Model size.}}
\label{tab:modelsize}
\vspace{-0.2cm}
\scriptsize
\resizebox{\textwidth}{!}{
\begin{tabular}{p{2.2cm}|cccccc}
\toprule
Model Size & NC$\uparrow$ & DAC$\uparrow$ & TTC$\uparrow$ & Comf.$\uparrow$ & EP$\uparrow$ & \cellcolor{gray!30}PDMS$\uparrow$ \\
\midrule
5B LoRA           & 92.4 & 88.0 & 91.0 & 99.9 & 67.5 & \cellcolor{gray!30}74.9 \\
14B LoRA          & 96.3 & 91.3 & 95.7 & 99.4 & 71.6 & \cellcolor{gray!30}80.6 \\
5B Full Fine-tune & \textbf{99.2} & \textbf{97.5} & \textbf{98.7} & \textbf{100} & \textbf{83.5} & \cellcolor{gray!30}\textbf{90.9} \\
\bottomrule
\end{tabular}
}
\end{minipage}

\end{table*}

\noindent \textbf{Why does \ourmethod generalize better in zero-shot setting?}
Our zero-shot visualizations in Fig.~\ref{fig:visual_compare} provide a clear explanation: \ourmethod maintains strong \emph{video--trajectory consistency} under domain shift, where the predicted trajectory aligns well with the imagined future scene evolution. In contrast, PWM often exhibits noticeable video--trajectory mismatch in zero-shot transfer, leading the planned motion to deviate from the future states implied by its imagination and resulting in error accumulation in recurrent rollout. For instance, in Fig.~\ref{fig:visual_compare}, PWM imagines a left-turning future, yet predicts a near-straight trajectory, revealing a clear inconsistency between its visual rollout and planned motion.

\begin{figure}[t]
    \centering
    \includegraphics[width=0.94\linewidth]{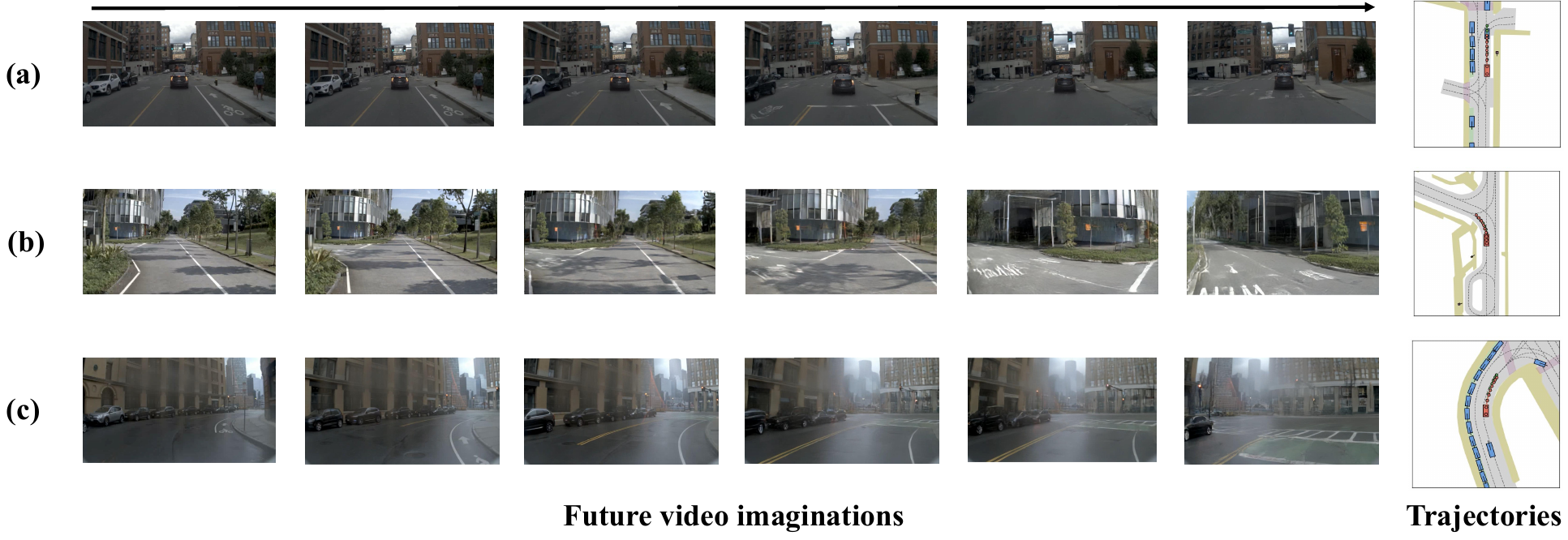}
    \caption{\textbf{Visualization of predicted video imaginations and corresponding trajectories.} 
The predicted trajectories follow the scene evolution in the generated future video frames, 
demonstrating strong video–trajectory consistency enabled by our unified generation framework.}
    \label{fig:visual1}
    \vspace{-0.55cm}
\end{figure}

\noindent\textbf{Simulation-Enhanced Real-World Transferring.}
\ourmethod further benefits from joint training with simulation data, as shown in Table~\ref{tab:ablation} (ID 2 vs. 4). Simulated environments such as CARLA provide diverse, high-quality corner-case scenarios that are hard to obtain at scale in real-world datasets, making them a valuable source of transferable driving priors. By jointly training on NAVSIM and simulation data, \ourmethod achieves improved planning performance in TTC (98.7) and PDMS (90.9) and maintains high performance in other planning metrics on real-world benchmarks, showing that simulation can enhance real-world planning. 


\subsection{Qualitative Analysis}
The qualitative results in Fig.~\ref{fig:visual1} further support the quantitative findings. Our \ourmethod generates future video imaginations that remain visually coherent over time while producing trajectories that stay well aligned with the evolving scene content. This strong video-trajectory consistency is a direct consequence of our unified generation design: instead of predicting visual futures and actions in separate stages, \ourmethod jointly decodes them within a shared latent generative process. As a result, the predicted trajectory more faithfully follows the semantic layout and motion trends implied by the generated future frames. \ourmethod not only achieves better planning metrics, but also produces more trustworthy future imaginations whose spatial evolution remains aligned with the intended driving behavior.

\subsection{Ablation Study}
\noindent
\textbf{Effect of key designs.}
Table~\ref{tab:ablation} shows that video supervision is the main driver of the gain: removing Video Loss drops PDMS from 90.9 to 71.4. Removing Video Continuation also degrades PDMS to 84.6, indicating that the video--action coupling should be preserved during rollout. CARLA mix training further improves PDMS from 90.5 to 90.9, showing the benefit of simulation data.

\noindent
\textbf{Sampling steps.}
Table~\ref{tab:step} shows that one sampling step is insufficient, while two steps already reach 90.9 PDMS and three steps bring no additional gain. This indicates that \ourmethod can perform efficient planning with very few flow-matching steps.

\noindent
\textbf{Prediction time horizon.}
Table~\ref{tab:futureframes} varies the video rollout length while fixing the trajectory horizon to $K{=}8$ (4s). Eight future frames perform best, whereas shorter rollouts insufficiently cover the action chunk and longer rollouts introduce additional drift.

\noindent
\textbf{Training strategy.}
Table~\ref{tab:trainstrategy} compares training from scratch, LoRA~\cite{hu2022lora}, and full fine-tuning. Full fine-tuning performs best, supporting that effective transfer requires adapting the video prior under joint video-level supervision. 

\noindent
\textbf{Model size.}
Table~\ref{tab:modelsize} further compares different model scales and tuning strategies, showing that full fine-tuning the 5B backbone performs best.

\section{Conclusion}
In this paper, we propose DriveVA, a unified video–action world model for autonomous driving. DriveVA jointly generates future video latents and trajectory tokens within a shared conditional generative process, improving video–trajectory consistency in real world rollout. Built on a large pretrained video generation backbone and further enhanced with progressive video continuation, DriveVA achieves state-of-the-art PDM-based performance on NAVSIM and shows strong zero-shot transfer to nuScenes and Bench2Drive without target-domain fine-tuning, which provides a new insight in this research domain.

\section*{Acknowledgements}
This work has been supported by the 
Centre for Spatial Intelligence (RCSI) at University of Bath,  the European Union under grant agreement no. 101136006-XTREME, and
the European Innovation Council under grant agreement no. 101257536-CEREBRIS.

%
%
\bibliographystyle{splncs04}
\bibliography{main}

\clearpage
\appendix
\section*{Appendix}
\section{Overview}
\label{sec:overview}

This supplementary material provides additional analyses for \ourmethod. We first describe the joint video--action formulation and the conditioning interface used by the flow-matching decoder. We then provide additional zero-shot qualitative examples, design ablations, a Bench2Drive Dev10 closed-loop study, and additional qualitative examples.

The supplementary material is organized as follows. Sec.~\ref{sec:joint} describes the joint video--action formulation and the conditioning and target-noising interface used in training. Sec.~\ref{sec:zeroshot} provides additional zero-shot visualizations, complementing the full zero-shot comparison reported in the main paper. Sec.~\ref{sec:consistency} provides DPVO-based external verification that the motion implied by the generated future videos is highly consistent with the trajectories predicted by \ourmethod. Sec.~\ref{sec:design_efficiency} analyzes the unified design and closed-loop behavior. Sec.~\ref{sec:failure} discusses representative failure cases. Sec.~\ref{sec:vis} presents more qualitative results.

\section{More Details about Joint Video--Action Modeling}
\label{sec:joint}

\paragraph{Joint prediction.}
Given the history observation buffer $\mathcal{O}_l$, language instruction $\mathcal{T}$, and current ego state $\mathbf{q}_l$, \ourmethod jointly predicts future video latents and future action tokens. We denote the conditioning context as $\mathbf{C}_l := (\mathcal{O}_l,\mathcal{T},\mathbf{q}_l)$ and write the joint rollout distribution as
\begin{equation}
\pi_{\theta}\!\left(\mathcal{F}_{l+1:l+N},\mathcal{A}_{l+1:l+K}\mid \mathcal{O}_l,\mathcal{T},\mathbf{q}_l\right).
\label{eq:joint_pred_main}
\end{equation}
This formulation can be interpreted as unifying future video continuation and IDM-style~\citep{du2023learning,zhou2024robodreamer} action grounding. From the chain rule, the same joint distribution can be written as
\begin{equation}
\resizebox{0.98\columnwidth}{!}{$
\underbrace{\pi_{\theta}\!\left(\mathcal{F}_{l+1:l+N},\mathcal{A}_{l+1:l+K}\mid \mathbf{C}_l\right)}_{\mathclap{\text{\ourmethod}}}
=
\underbrace{\pi_{\theta}\!\left(\mathcal{F}_{l+1:l+N}\mid \mathbf{C}_l\right)}_{\mathclap{\text{video continuation}}}
\underbrace{\pi_{\theta}\!\left(\mathcal{A}_{l+1:l+K}\mid \mathbf{C}_l,\mathcal{F}_{l+1:l+N}\right)}_{\mathclap{\text{action grounding}}}.
$}
\label{eq:joint_pred_decomp}
\end{equation}
The second factor reflects the intuition that, once a future visual evolution is specified, the future action chunk should be compatible with that visual evolution under the same driving context. Different from a cascaded implementation that first predicts a video and then applies a separate inverse-dynamics model, \ourmethod jointly denoises future video latents and action tokens in a single DiT. Self-attention therefore allows both modalities to exchange information during generation, making the trajectory an action grounding of the same future scene evolution represented by the generated video latents.

\paragraph{Conditioning interface.}
In training, the input sequence is split into a fixed condition block and a generative target block. The condition block contains history visual latents and ego-state tokens; it is kept noise-free throughout the flow-matching process. The generative target block contains future video latent tokens and future action tokens; Gaussian noise is applied only to this target block, and the DiT predicts the target velocity for denoising. This design has two practical advantages. First, it avoids leakage from future tokens into the condition stream, since only the target tokens are corrupted and reconstructed. Second, it keeps the training and inference interfaces consistent: at inference time the model receives the same type of history and ego-state condition tokens, and generates the future visual and action targets from noise. In the default setting, future video and action target tokens attend to each other during denoising, which allows bidirectional coupling between imagined scene evolution and planned ego motion.

\section{Additional Zero-Shot Results}
\label{sec:zeroshot}

The main paper reports the full zero-shot comparison on nuScenes and Bench2Drive, including both DriveVLA-W0 and PWM under the same NAVSIM-training protocol. In this section, we provide additional qualitative examples and extra unseen-dataset checks to complement the main quantitative results.

\begin{figure}[t]
    \centering
    \includegraphics[width=1.0\linewidth]{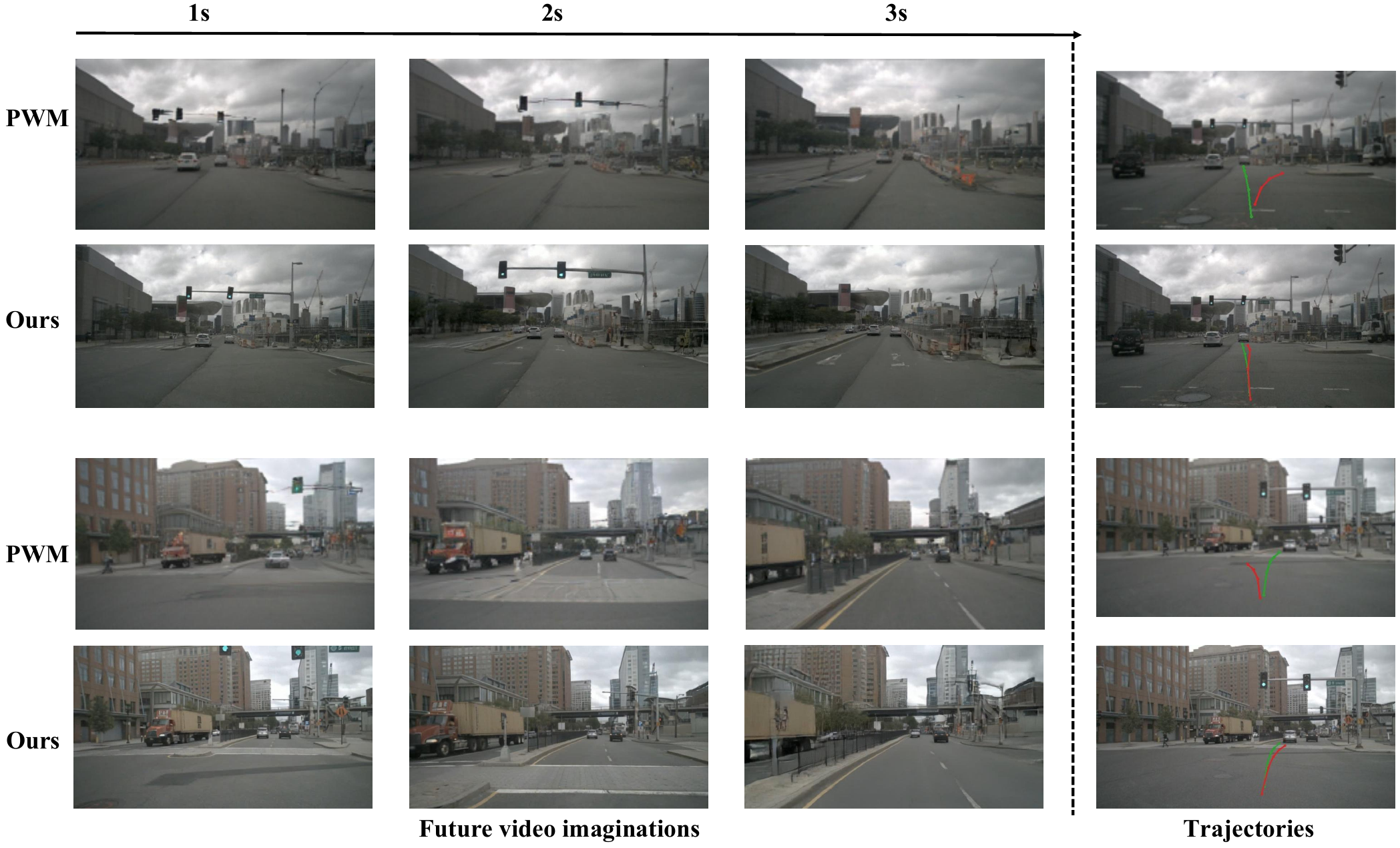}
    \caption{\textbf{Additional comparison of video--trajectory consistency in zero-shot unseen nuScenes scenarios.} We compare PWM~\citep{zhao2025forecasting} and \ourmethod on representative unseen nuScenes cases. For each case, we show imagined future video frames at 1s, 2s, and 3s, together with the predicted trajectory. PWM exhibits mismatched turning behavior between its visual rollout and planned ego motion. In contrast, \ourmethod maintains stronger alignment between future video imagination and trajectory prediction. Here, \textcolor{green}{green} denotes the GT trajectory and \textcolor{red}{red} the predicted trajectory.}
    \label{fig:visual_compare_supp}
\end{figure}

Fig.~\ref{fig:visual_compare_supp} provides additional qualitative evidence under zero-shot transfer. PWM can produce future videos and trajectories that imply different maneuvers, while \ourmethod keeps the generated future and planned trajectory better aligned. This suggests that the transfer gain is not only due to better endpoint accuracy, but also to tighter coupling between visual forecasting and action generation.

\section{DPVO-based External Verification of Video--Trajectory Consistency}
\label{sec:consistency}

To further verify that the predicted future videos and trajectories are mutually consistent, we conduct both quantitative and qualitative analysis using visual-odometry-based trajectory reconstruction.
Specifically, for each sample, we run DPVO~\cite{teed2023deep} on (i) the ground-truth future video clip and (ii) the future video clip generated by \ourmethod, and obtain the corresponding reconstructed camera trajectories.
We then compare the reconstructed trajectory from the ground-truth future video with the ground-truth ego trajectory, and compare the reconstructed trajectory from the predicted future video with the trajectory predicted by \ourmethod.

Since monocular visual odometry is inherently ambiguous up to scale, we first align each reconstructed trajectory to its corresponding reference trajectory using a 2D similarity transform before computing the trajectory error.
Let $\tau=\{\mathbf{p}_t\}_{t=1}^{T}$ denote a reference trajectory and $\hat{\tau}=\{\hat{\mathbf{p}}_t\}_{t=1}^{T}$ denote the aligned reconstructed trajectory, where $\mathbf{p}_t,\hat{\mathbf{p}}_t \in \mathbb{R}^2$ are ego positions in the ground plane.
We measure their discrepancy by the average L2 error over the future 4s horizon:
\begin{equation}
\mathrm{Avg.\ L2}(\hat{\tau},\tau)=
\frac{1}{T}\sum_{t=1}^{T}\left\|\hat{\mathbf{p}}_t-\mathbf{p}_t\right\|_2.
\label{eq:avg_l2_consistency}
\end{equation}

Table~4 in main paper reports the quantitative results.
The errors are consistently small for both the \emph{GT trajectory vs.\ GT-video reconstruction} pair and the \emph{predicted trajectory vs.\ predicted-video reconstruction} pair.
On NAVSIM, the average L2 error is 0.09 for ground-truth video reconstruction and 0.16 for predicted video reconstruction.
On zero-shot nuScenes, the corresponding errors are 0.07 and 0.14, respectively.
Averaged across the two benchmarks, the reconstruction error remains as low as 0.08 for the ground-truth branch and 0.15 for the predicted branch.

These results indicate that the motion implied by the visual evolution in the video is highly consistent with the corresponding trajectory sequence.
In particular, the low error on the predicted branch shows that the trajectory generated by \ourmethod is not only plausible as a planning output, but is also faithfully supported by the motion cues embedded in the generated future video.
We further provide qualitative visualizations showing that the reconstructed trajectories closely follow the corresponding ground-truth and predicted trajectories, which is consistent with the quantitative results.
Together, these findings provide direct evidence that \ourmethod achieves strong video--trajectory consistency by jointly decoding future videos and actions within a unified generative process.

\begin{figure}[t]
    \centering
    \includegraphics[width=1.0\linewidth]{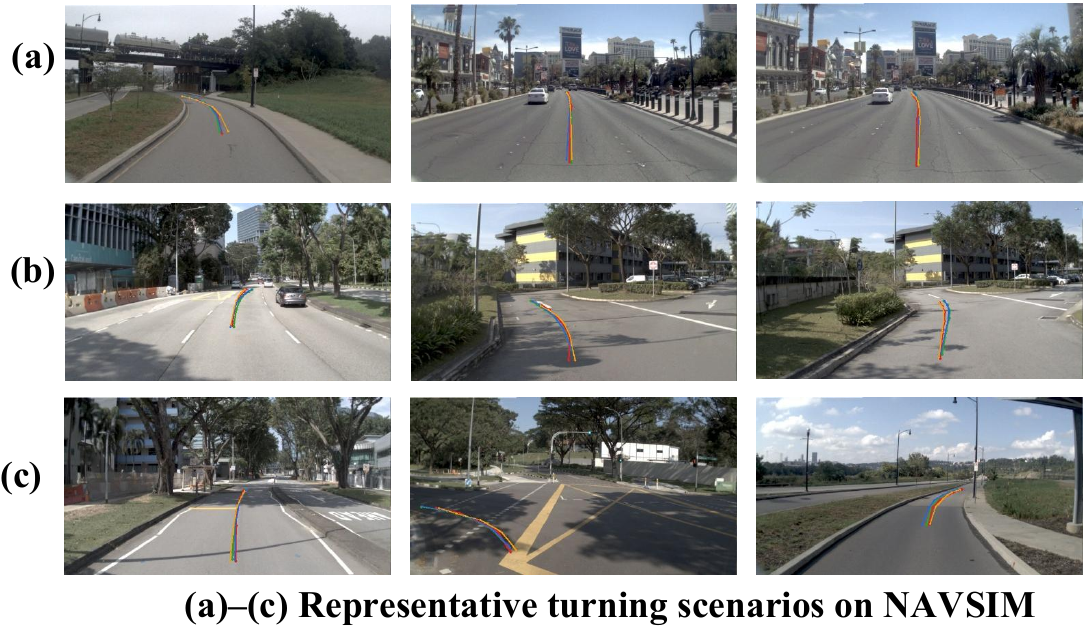}
    \caption{\textbf{DPVO-based qualitative analysis of video--trajectory consistency on NAVSIM.}
    We visualize representative turning scenarios and compare the reference trajectories with DPVO-reconstructed trajectories from ground-truth and predicted future videos.
    \textcolor{green}{GT Future} and \textcolor{red}{Pred Future} indicate the ground-truth and predicted trajectories, respectively.
    \textcolor{dpvogat}{DPVO(gt img)} and \textcolor{dpvopred}{DPVO(pred img)} indicate the trajectories reconstructed by DPVO from the ground-truth and predicted future videos, respectively.
    The strong overlap among these curves across diverse turning cases shows that the motion implied by the generated future video is highly consistent with the predicted trajectory.}
    \label{fig:visual_dpvo_navsim}
    \vspace{-0.4cm}
\end{figure}

The qualitative visualizations in Fig.4 in main paper and Fig.~\ref{fig:visual_dpvo_navsim} further make this consistency explicit. Across lane-change, right-turn, straight-driving, and turning scenarios, the trajectory reconstructed from the generated future video, \textcolor{dpvopred}{DPVO(pred img)}, closely overlaps with the model-predicted trajectory, \textcolor{red}{Pred Future}, throughout the rollout. This means that an independent visual odometry system can recover essentially the same ego motion directly from the generated frames, providing strong evidence that \ourmethod achieves excellent consistency between predicted videos and predicted trajectories. Together, these findings show that the unified generative process in \ourmethod couples visual imagination and planning behavior in a geometrically meaningful way.

\section{Design and Closed-Loop Analysis}
\label{sec:design_efficiency}
\subsection{Preliminary closed-loop evaluation on Bench2Drive Dev10}

\begin{table}[H]
\centering
\caption{\textbf{Preliminary closed-loop performance on the Bench2Drive Dev10 split.} DS and SR denote Driving Score and Success Rate. Dev10 is a small diagnostic split, so this result is treated as auxiliary closed-loop evidence rather than a substitute for full Bench2Drive evaluation.}
\label{tab:bench2drive_dev10}
\setlength{\tabcolsep}{8pt}
\begin{tabular}{l|cc}
\toprule
Method & DS $\uparrow$ & SR $\uparrow$ \\
\midrule
DriveTransformer~\citep{jia2025drivetransformer} & 60.45 & 30.00 \\
DriveMamba-12L~\citep{su2026drivemamba} & 66.50 & 40.00 \\
\textbf{Ours} & \textbf{67.47} & \textbf{70.00} \\
\bottomrule
\end{tabular}
\end{table}

Table~\ref{tab:bench2drive_dev10} reports a preliminary closed-loop study on Bench2Drive Dev10. \ourmethod improves the success rate over both baselines and obtains a competitive driving score, suggesting promising transfer from NAVSIM-trained video--action modeling to closed-loop simulation. Since Dev10 contains only a limited number of routes, we treat it as an auxiliary check rather than a replacement for full Bench2Drive evaluation.

\subsection{Additional design ablations}

\begin{table*}[t]
\centering
\caption{\textbf{Additional ablation studies on masking strategy and dual prediction.} All results use NAVSIM PDM-based metrics.}
\label{tab:ablation_misc}
\scriptsize
\setlength{\tabcolsep}{5pt}
\resizebox{\textwidth}{!}{
\begin{tabular}{l|l|cccccc}
\toprule
Module & Setting Variant & NC$\uparrow$ & DAC$\uparrow$ & TTC$\uparrow$ & Comf.$\uparrow$ & EP$\uparrow$ & \cellcolor{gray!30}PDMS$\uparrow$ \\
\midrule
\multicolumn{8}{c}{\textbf{Masking Strategy}} \\
\midrule
Masking Strategy & Causal Mask  & 99.0 & 97.1 & 98.2 & 99.8 & 82.5 & \cellcolor{gray!30}90.1 \\
Masking Strategy & Bidirectional  & \textbf{99.2} & \textbf{97.5} & \textbf{98.7} & \textbf{100} & \textbf{83.5} & \cellcolor{gray!30}\textbf{90.9} \\
\midrule
\multicolumn{8}{c}{\textbf{Prediction Target}} \\
\midrule
Prediction Target & Action Only & 89.7 & 87.4 & 89.9 & 34.9 & 27.3 & \cellcolor{gray!30}47.0 \\
Prediction Target & Video + Action & \textbf{99.2} & \textbf{97.5} & \textbf{98.7} & \textbf{100} & \textbf{83.5} & \cellcolor{gray!30}\textbf{90.9} \\
\bottomrule
\end{tabular}}
\end{table*}

Table~\ref{tab:ablation_misc} provides additional analysis beyond the main ablation table. The causal-mask variant restricts the interaction between future video and action tokens, reducing PDMS from 90.9 to 90.1. This suggests that mutual interaction during denoising is helpful, even though the condition tokens remain fixed and noise-free. The action-only variant performs much worse than the default video-plus-action target, confirming that future-video prediction is not merely an auxiliary visualization output; it provides dense temporal grounding for the action tokens.


\section{Failure Cases and Limitations}
\label{sec:failure}

\begin{figure}[t]
    \centering
    \includegraphics[width=1.0\linewidth]{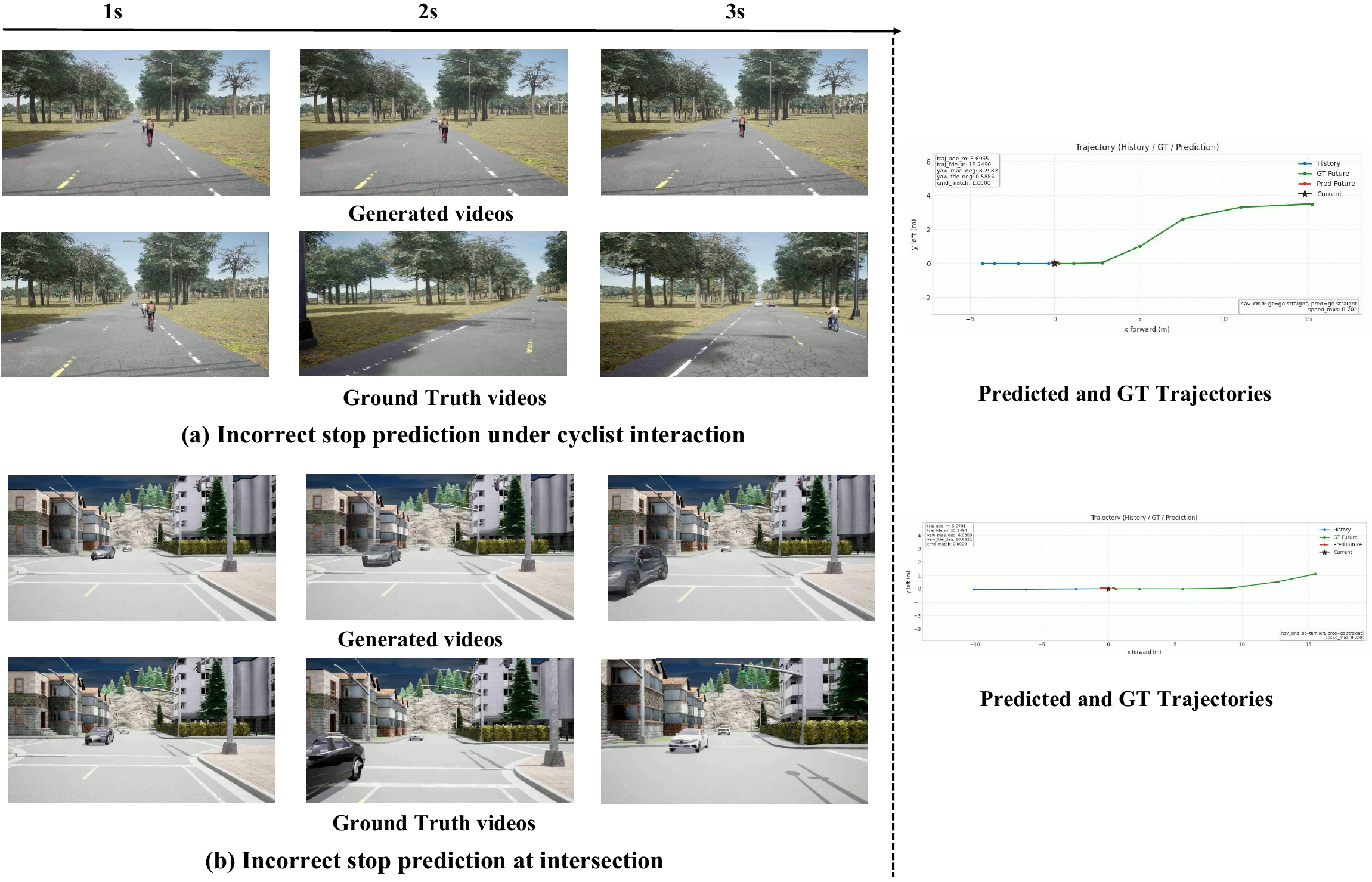}
    \caption{\textbf{Failure cases with consistent but incorrect video--trajectory prediction.} We compare generated future videos and predicted trajectories with ground-truth future videos and trajectories. In both examples, the predicted future mode differs from the ground truth, and the predicted trajectory follows the generated video. Here, \textcolor{green}{green} denotes the GT trajectory and \textcolor{red}{red} the predicted trajectory.}
    \label{fig:visual_supp_badcase}
\end{figure}

Fig.~\ref{fig:visual_supp_badcase} illustrates two representative failure cases. In the first case, the ground-truth ego vehicle bypasses the cyclist, whereas \ourmethod predicts a more conservative stopping behavior. In the second case, the ground-truth ego vehicle proceeds through the intersection, while \ourmethod again predicts stopping and outputs a nearly stationary trajectory. These examples show that the video and trajectory branches remain aligned even when the selected future mode is wrong. In other words, low-level visual artifacts are not the only possible source of planning error; causal or intention-level mistakes in future imagination can also lead to consistent but incorrect trajectories. Improving causal scene understanding and multi-modal future reasoning is therefore an important direction for future work.


\section{More Qualitative Results}
\label{sec:vis}

\subsection{Additional predicted videos and trajectories}

\begin{figure}[t]
    \centering
    \includegraphics[width=1.0\linewidth]{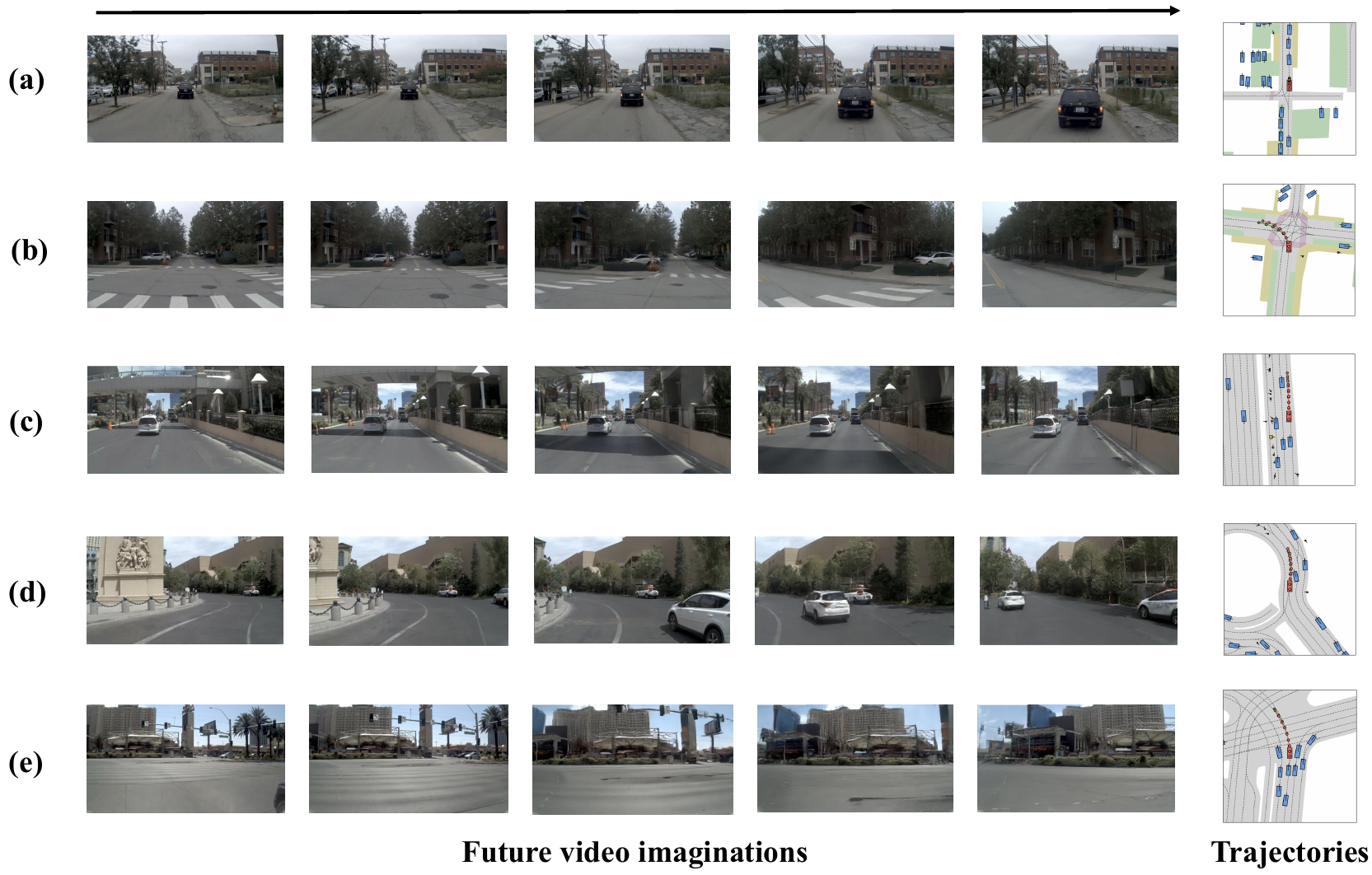}
    \caption{\textbf{More visualizations of predicted video imaginations and corresponding trajectories.} The predicted trajectories follow the scene evolution in the generated future video frames, demonstrating video--trajectory consistency enabled by the unified generation framework. Here, \textcolor{green}{green} denotes the GT trajectory and \textcolor{red}{red} the predicted trajectory.}
    \label{fig:visual1_supp}
\end{figure}

Following the qualitative analysis in the main paper, Fig.~\ref{fig:visual1_supp} provides more examples of generated future videos and their corresponding trajectories. These examples show that the predicted trajectories evolve consistently with the generated future frames, supporting the role of the unified video--action rollout in maintaining alignment between visual imagination and ego motion.

\subsection{Additional zero-shot visualizations}

\begin{figure}[t]
    \centering
    \includegraphics[width=1.0\linewidth]{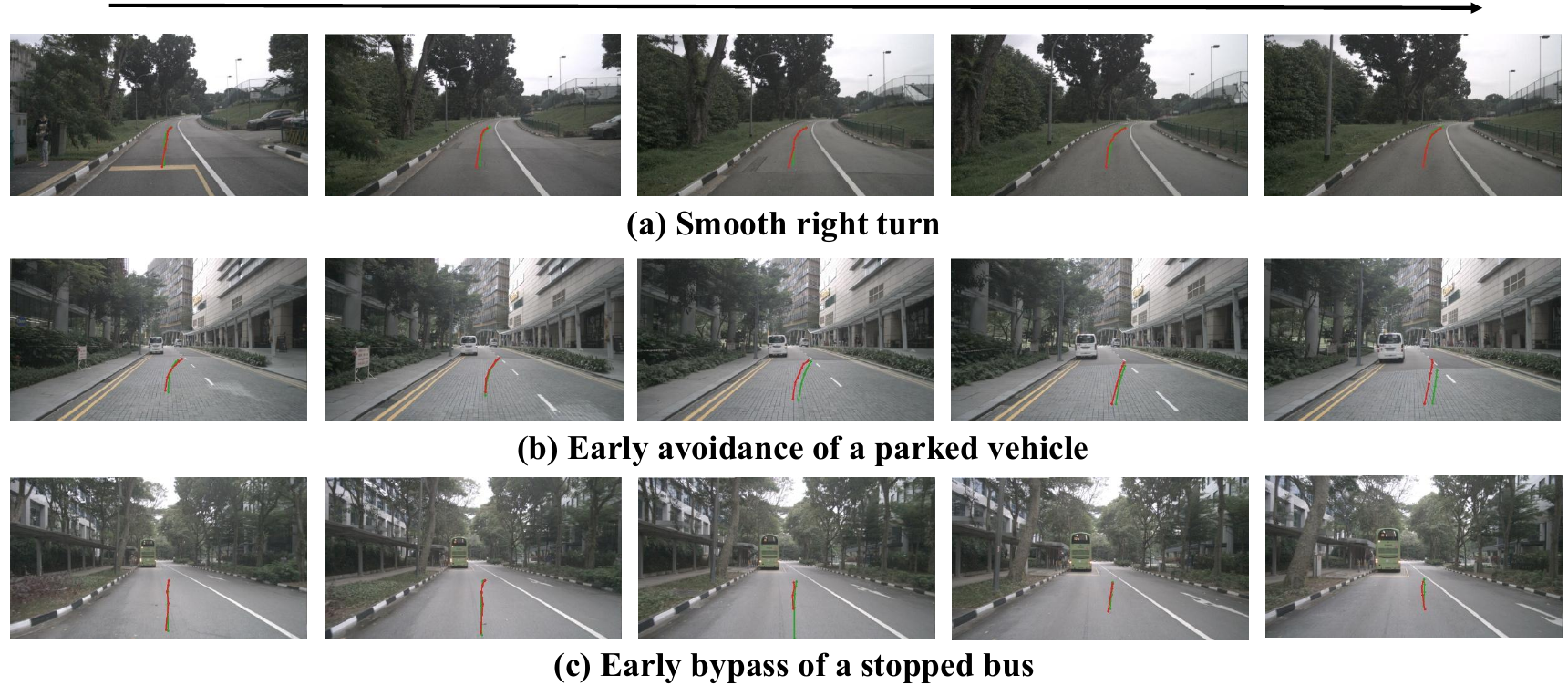}
    \caption{\textbf{Temporal video--trajectory consistency in zero-shot unseen nuScenes scenarios.} From left to right, the generated future video frames and overlaid trajectories evolve consistently over time in right-turn, parked-vehicle avoidance, and stopped-bus bypass scenarios. Here, \textcolor{green}{green} denotes the GT trajectory and \textcolor{red}{red} the predicted trajectory.}
    \label{fig:visual_temporal_supp}
\end{figure}

\begin{figure}[t]
    \centering
    \includegraphics[width=1.0\linewidth]{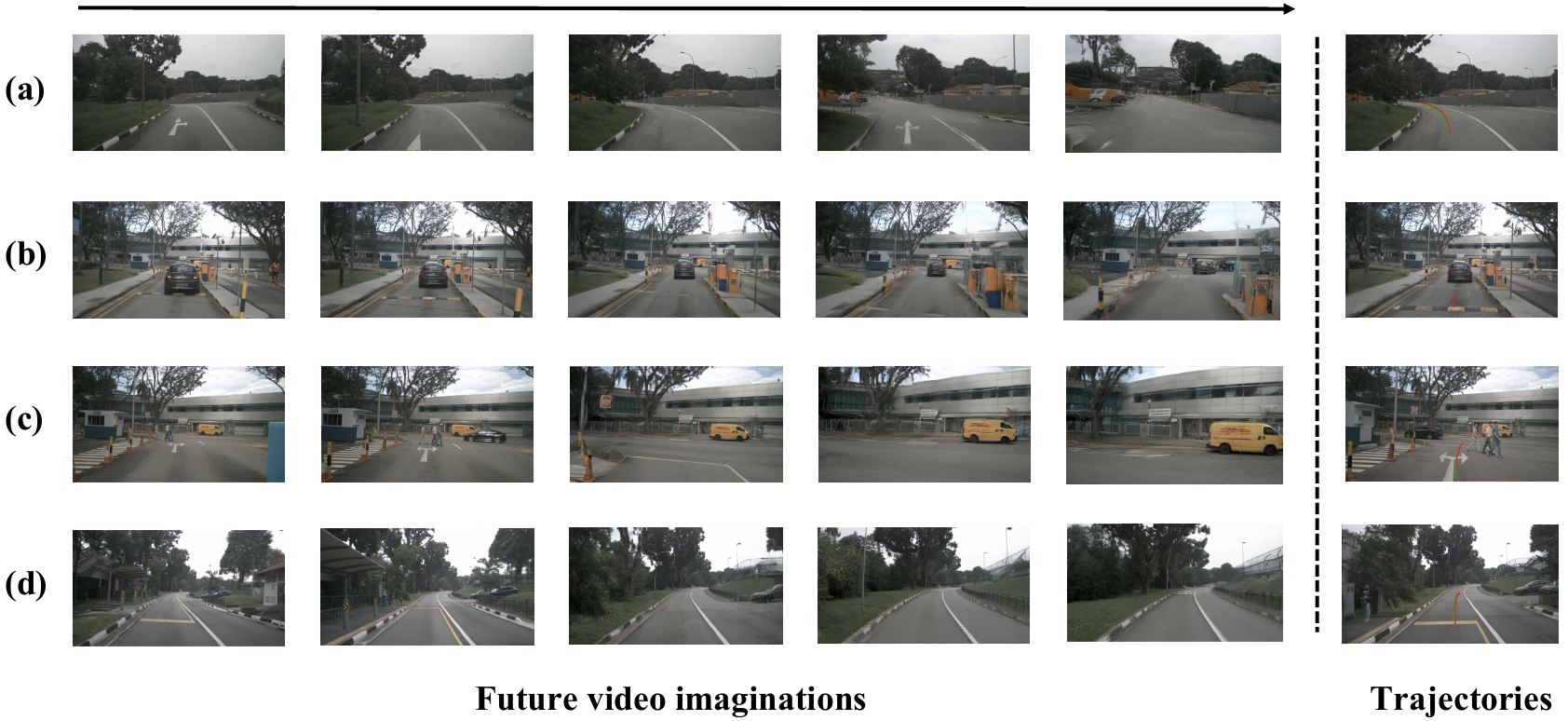}
    \caption{\textbf{Zero-shot generalization on nuScenes with predicted video imaginations and corresponding trajectories.} The trajectories remain aligned with the generated video evolution under dataset shift. Here, \textcolor{green}{green} denotes the GT trajectory and \textcolor{red}{red} the predicted trajectory.}
    \label{fig:visual2}
\end{figure}

\begin{figure}[t]
    \centering
    \includegraphics[width=1.0\linewidth]{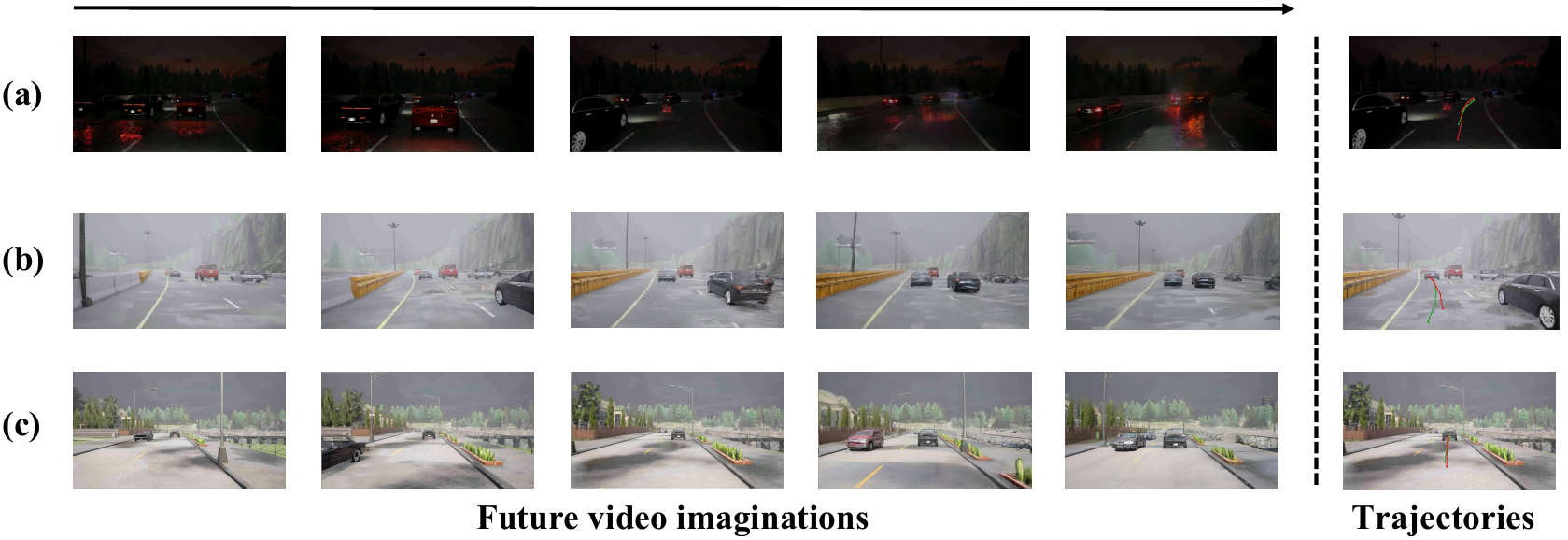}
    \caption{\textbf{Zero-shot generalization on CARLA with predicted video imaginations and corresponding trajectories.} The trajectories remain aligned with the generated video evolution under real-to-simulation domain shift. Here, \textcolor{green}{green} denotes the GT trajectory and \textcolor{red}{red} the predicted trajectory.}
    \label{fig:visual3}
\end{figure}

Figs.~\ref{fig:visual_temporal_supp}, \ref{fig:visual2}, and~\ref{fig:visual3} show additional zero-shot examples on unseen nuScenes and CARLA scenarios. Although the visual domains differ from NAVSIM, the generated future videos and predicted trajectories remain aligned in turning, bypassing, and straight-driving scenarios. These results complement the quantitative transfer results by showing how the model behaves at the level of visual rollout.

\end{document}